\pdfoutput=1
\documentclass{article}

\PassOptionsToPackage{dvipsnames}{xcolor}
\usepackage{iclr2027_conference,times}
\iclrfinalcopy

\usepackage[utf8]{inputenc}
\usepackage[T1]{fontenc}
\usepackage{url}
\usepackage{microtype}
\usepackage{nicefrac}

\usepackage{amsmath,amsfonts,amssymb}
\usepackage{algorithmic}
\usepackage{algorithm}
\usepackage{array}
\usepackage{textcomp}
\usepackage{verbatim}
\usepackage{graphicx}
\usepackage{booktabs}
\usepackage{multirow}
\usepackage{caption}
\usepackage{subcaption}
\usepackage{xcolor}
\usepackage{placeins}

\usepackage{titlesec}
\titlespacing*{\section}{0pt}{6pt plus 1pt minus 1pt}{3pt plus 1pt minus 1pt}
\titlespacing*{\subsection}{0pt}{4pt plus 1pt minus 1pt}{2pt plus 1pt minus 1pt}
\titlespacing*{\subsubsection}{0pt}{3pt plus 1pt minus 1pt}{1pt plus 1pt minus 1pt}
\titlespacing*{\paragraph}{0pt}{2pt plus 1pt minus 1pt}{1em}
\usepackage{hyperref}

\title{CST-WM: A Causally Structured World Model for Embodied Visual Tracking}

\author{Junyi Hu, Shuaihang Yuan, Jiazhao Liang, Yi Fang\thanks{Corresponding author.} \\
New York University Abu Dhabi \\
\texttt{jh10472@nyu.edu}}

\begin{document}

\maketitle
\lhead{Preprint}

\begin{abstract}

Embodied visual tracking requires a robot to choose actions that keep a moving target observable at a suitable distance, and to recover it after occlusion, out-of-view drift, or distractor crossings. We cast the task as planning over future target evidence with an action-conditioned world model. In logged tracking data, however, the behavior policy's actions are correlated with where the target is, so a generic predictor can learn a shortcut: it writes the current action directly into its prediction of target evidence, instead of letting the action affect that evidence only by moving the robot and changing what it observes. We call this failure causal hallucination; the resulting rollouts look plausible but rank candidate actions for the wrong reason. We propose CST-WM, a causally structured world model whose state is split into target-evidence, robot, and observation branches. Its transition removes the same-step edge from action to target evidence but keeps the path through robot motion and the resulting views, so candidate actions are still distinguished by their predicted ego-motion. With rollout-based model-predictive control, a single model handles both steady following and re-acquisition after target loss. On EVT-Bench and Habitat 3.0, covering standard tracking, target-loss recovery, and cross-dataset transfer, CST-WM improves following, distance-range control, safety, and re-acquisition over reactive trackers and world-model baselines, and removing the action mask causes the largest drop in re-acquisition among our ablations. Offline, CST-WM has lower multi-step rollout error, and its ranking of candidate actions agrees better with the simulator's. On a Unitree Go2 quadruped, CST-WM succeeds in 20 of 30 real-world trials under occlusion, distractor crossing, and fast motion, against 14 for TrackVLA~\citep{trackvla}.
Project page: \url{https://junyi2005.github.io/cst-wm/}

\end{abstract}

\section{Introduction}
\label{sec:intro}

Embodied visual tracking~\citep{luo2018e2eat, advat, evt, trackvla} appears simple: a robot only needs to stay with a moving target. In practice, reliable following requires the robot to preserve target observability, regulate following distance, and remain safe while both agents move through cluttered, partially observable environments~\citep{followanything, rspt, oclrpf}. The hardest moments are not when the target is clearly visible, but when it is temporarily occluded, leaves the field of view, or becomes confusable with distractors~\citep{DistractionRobustAVT, trackvlapp, li2020poseassisted}. The value of an action therefore depends not only on how it changes the robot state, but also on how it changes future target visibility and apparent scale. Embodied visual tracking is thus a predictive decision problem rather than a reactive mapping from the current frame to the next control~\citep{ha2018worldmodels, hafner2019planet, nwm}. Learning the predictor from logged tracking data has a pitfall: the recorded robot turns toward and slows down for the target, so its actions are correlated with target evidence. We call this failure mode causal hallucination: an action-conditioned predictor can exploit that correlation by attributing to the current action a direct effect on target evidence, instead of letting the action influence it only through robot motion and the resulting change of view.

Neither dominant formulation covers these moments. Reactive trackers~\citep{ActiveTracking, evt, trackvla, sda, advat, luo2018e2eat, uninavid, openvla, pi0} are myopic: once the target disappears or becomes ambiguous, they offer little basis for choosing actions that support multi-step recovery. A generic action-conditioned world model~\citep{nwm, hafner2020dreamer, hafner2023dreamerv3, bruce2024genie} is not sufficient either: a predictor that writes the current action directly into its target-related state produces plausible futures with the wrong semantics for re-acquisition. The error is invisible on logged action sequences, where both explanations predict similar futures, and surfaces where planning relies on the model, under candidate actions that the recording robot never took.

\begin{figure*}[t]
  \centering
  \includegraphics[width=\textwidth]{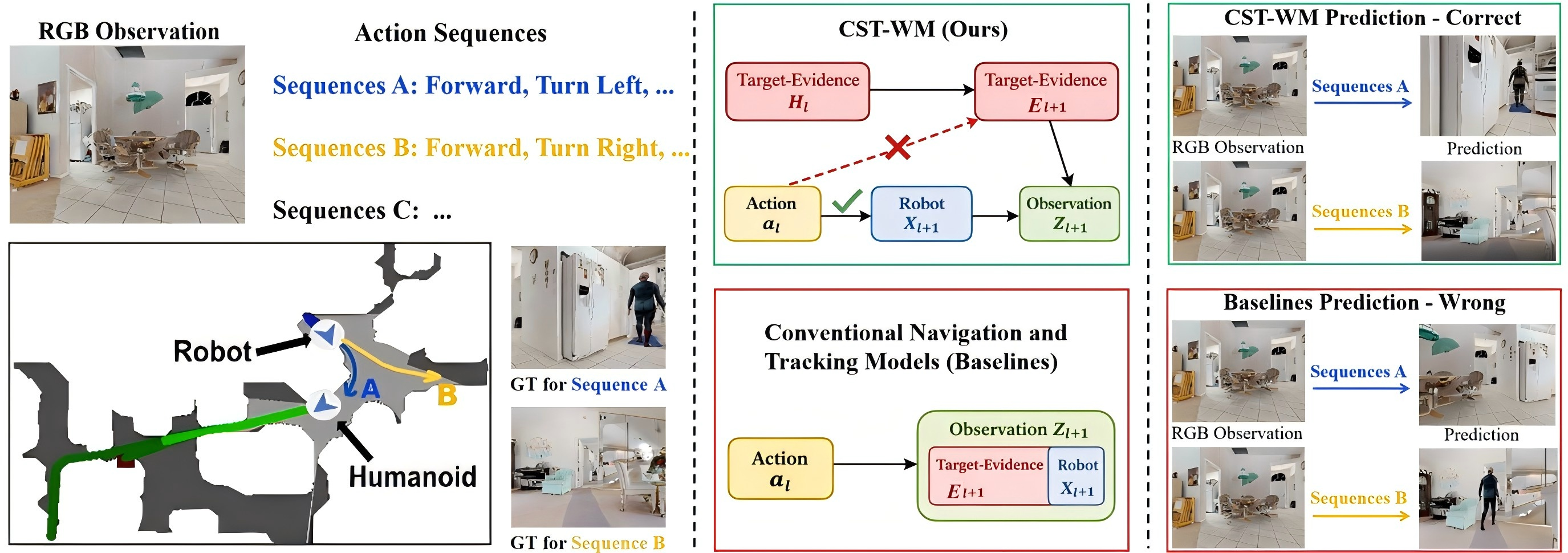}
  \caption{Causal structure of action-conditioned prediction in embodied visual tracking. Given an egocentric observation and candidate action sequences (A: forward, turn-left; B: forward, turn-right), the model must imagine which sequence preserves target observability and following distance. Conventional world models route the current action directly into the target-evidence update, producing plausible rollouts that fail to discriminate A from B. CST-WM removes this direct edge: the target-evidence branch $E_{\ell+1}$ is updated without action injection, and the action only reaches future observations through the robot branch $R_{\ell+1}$. The resulting predictions distinguish the two sequences and agree with the simulator.}
  \label{fig:intro}
\end{figure*}

As illustrated in Figure~\ref{fig:intro}, tracking does not require recovering the target’s full state at every step: for planning, a compact target evidence that says whether the target is observable and whether its apparent scale is compatible with a valid following distance suffices. We propose CST-WM, a causally structured world model with a target-evidence branch, a robot branch, and an observation branch. Its transition updates the target-evidence branch without the current action, which enters only the robot branch and reaches future observations through that route, and the target evidence of an imagined future is read from the observation branch by an evidence head distilled from the detector. Different candidate actions thus change later target evidence through the views they produce; while the target is hidden, candidates are ranked by whether their imagined views bring it back, so one rollout-based planner supports both stable following and re-acquisition.

We summarize our contributions as follows: (i)~we formulate embodied visual tracking as planning over future target evidence and propose CST-WM, a causally structured world model that separates target evidence, robot motion, and observation dynamics while blocking direct action injection into the target-evidence branch; (ii)~we introduce a compact target-evidence representation of observability and apparent scale for following, recovery, and safety-aware planning, read on imagined futures by an evidence head distilled from an open-vocabulary detector, with no privileged geometric supervision at test time; and (iii)~on EVT-Bench single-target tracking~\citep{trackvla} and Habitat 3.0~\citep{h3}, against reactive trackers and an adapted NWM planner that differs only in the transition, CST-WM improves following, distance-range control, safety, and re-acquisition, and its imagined futures have higher rollout fidelity and rank counterfactual candidates closer to the simulator, with structural diagnostics, planning cost, and real-robot trials.

\section{Related Work}
\label{sec:related_work}
\subsection{Embodied visual tracking}
Embodied visual tracking requires a robot to follow a moving target from egocentric observations while preserving visibility, regulating following distance, and remaining safe~\citep{oclrpf, rpfsearch, socialnav_eval}. Detection-plus-planning systems achieve basic following but degrade when the target becomes occluded or visually ambiguous in cluttered scenes. A major recent direction is reactive policy learning: reinforcement-learning-based active trackers~\citep{ActiveTracking, EvasionPolicies, DistractionRobustAVT, poli}, offline-RL extensions such as EVT~\citep{evt, offline, Liu2024TrajectorySamplingOfflineRL}, and occlusion-aware trackers with explicit re-acquisition modes~\citep{sun2026oavat}, alongside social-aware and vision-language-action variants that map rich visual context to control~\citep{sda, wu2025vlmgoodassistantenhancing, app15041907, liu2025adaptive, uninavid, trackvla, trackvlapp}. These methods strengthen robustness under occlusion and distractors, but they still mainly map recent observations to actions rather than comparing how alternative actions affect future target observability and scale over multiple steps.
\subsection{World models}
World models instead learn latent dynamics from observation histories and support planning by rolling out futures before action execution. From early latent-dynamics methods~\citep{ha2018worldmodels, hafner2019planet} to the Dreamer family and its embodied extensions~\citep{hafner2020dreamer, hafner2020dreamerv2, hafner2023dreamerv3, wu2023daydreamer}, and to broader generative or driving environments~\citep{bruce2024genie, wang2023drivedreamer, wang2024drivewm, Li2025DriVerse, Hu2023GAIA1}, this paradigm enables comparing multiple candidate futures before acting. The most directly relevant prior work in embodied navigation is Navigation World Models~\citep{nwm}, which shows that diffusion-based world modeling can support egocentric planning. WoMAP~\citep{yin2025womap} distills open-vocabulary detector confidences into a reward head of a latent world model for active object localization, so imagined states are scored without decoding them. CST-WM reads imagined target evidence the same way and additionally constrains how the action may reach it. A generic action-conditioned transition can write the current action directly into its target-evidence branch, a tracking-specific form of causal hallucination related to shortcut learning~\citep{geirhos2020shortcut} and to causal confusion in imitation learning~\citep{dehaan2019causal}, where a policy relies on features correlated with the expert's actions; here the confounded quantity is the model's prediction of target evidence, which planning queries under actions the data never contain. CST-WM therefore excludes the current action from the target-evidence branch while it still drives the robot and observation branches.

\section{Method}

\subsection{Problem Definition}

\begin{figure*}[t]
  \centering
  \includegraphics[width=\textwidth]{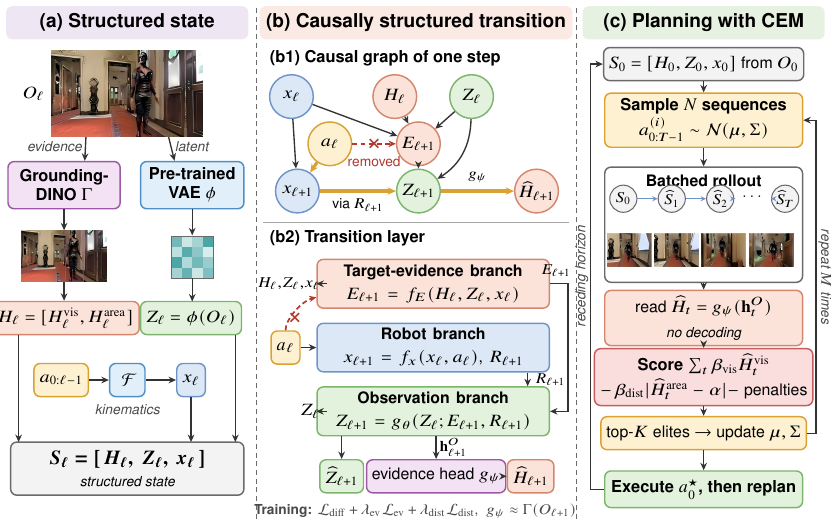}
  \caption{Overview of CST-WM. (a)~\textit{State.} GroundingDINO reads the target evidence $H_{\ell}$ (confidence, box area) from $O_{\ell}$, a frozen VAE encodes $O_{\ell}$ into $Z_{\ell}$, and the pose $x_{\ell}$ is integrated from past actions. (b)~\textit{Transition.} (b1)~The edge $a_{\ell} \rightarrow E_{\ell+1}$ is removed, and the action reaches the next target evidence only along the highlighted path through $R_{\ell+1}$ and $Z_{\ell+1}$. (b2)~Each layer updates the target-evidence, robot, and observation branches in this order, and the evidence head $g_{\psi}$, distilled from the detector, reads $\widehat{H}_{\ell+1}$ from the observation branch. (c)~\textit{Planning.} CEM rolls out $N$ sampled sequences, scores the evidence read from the imagined views, refits to the top $K$, and executes the best first action.}
  \label{fig:pipeline}
\end{figure*}

At each step $\ell$ the robot receives an egocentric RGB observation $O_{\ell} \in \mathbb{R}^{h \times w \times 3}$ and executes an action $a_{\ell}=(v_{\ell}, \omega_{\ell}) \in \mathcal{A} \subseteq \mathbb{R}^{3}$, with planar linear velocity $v_{\ell} \in \mathbb{R}^{2}$ and angular velocity $\omega_{\ell} \in \mathbb{R}$. The robot never observes the target's state directly, and an episode succeeds if it keeps facing the moving target at a following distance of 1--3\,m. Because the target moves on its own and the scene can hide it, an action is judged mainly by what the robot sees afterwards: whether the target stays in view, and how large it appears. We therefore predict these consequences with a world model and choose actions by planning over them (Figure~\ref{fig:pipeline}).

\subsection{Causally Structured World Model}
\label{sec:causal-diffusion}

\subsubsection{Structured State}
The planner needs to know whether the target is visible and how far away it is, and both can be read from the image: GroundingDINO~\citep{dino} returns a confidence $H^{\mathrm{vis}}_{\ell} \in [0,1]$ and a box whose relative area gives the apparent scale $H^{\mathrm{area}}_{\ell} \in [0,1]$:
\begin{equation}
\label{eq:H-score}
H_{\ell} = \Gamma(O_{\ell}) = [\, H^{\mathrm{vis}}_{\ell},\, H^{\mathrm{area}}_{\ell}\,] \in [0,1]^{2}.
\end{equation}
The score varies monotonically with distance over most of the following range and saturates only at very close range (Figure~\ref{fig:grounddino} in Appendix~\ref{app:signal}), so $H_{\ell}$ serves as an observability and distance signal without privileged geometry. The rest of the scene enters through the latent $Z_{\ell} = \phi(O_{\ell})$ of a pre-trained VAE~\citep{svd} with decoder $\mathcal{D}$, and the robot pose $x_{\ell} \in \mathbb{R}^{3}$ (planar position and yaw) is integrated from past actions, $x_{\ell} = \mathcal{F}(a_{0:\ell-1})$. The state is $S_{\ell} = [H_{\ell}, Z_{\ell}, x_{\ell}]$; the transition receives $H_{\ell}$ as the scalar $0.4\,H^{\mathrm{vis}}_{\ell} + 0.6\,(H^{\mathrm{area}}_{\ell})^{1/2}$ and conditions on the current step only, without observation history, while the planner uses the two components read by the evidence head.

\subsubsection{Where the Action May Enter}
Physically, the action can change target evidence only by moving the camera: $a_{\ell} \rightarrow x_{\ell+1} \rightarrow Z_{\ell+1} \rightarrow H_{\ell+1}$. Logged data offer a second route. The behavior policy that records a tracking dataset turns toward a target drifting out of view and slows down near it, so the current action and the next target evidence share a common cause, the target's position. A transition whose target representation reads $a_{\ell}$ can fit this correlation and learn what the behavior policy usually sees after an action rather than what the action does to the view; this is causal hallucination (Section~\ref{sec:intro}). On logged actions both routes predict similar futures, so training hardly separates them, but planning scores counterfactual candidates that the behavior policy never chose.

We therefore give the action only the physical route, with two requirements:
\begin{equation}
\label{eq:readout}
\widehat{H}_{\ell+1} = g_{\psi}\bigl(\mathbf{h}^{O}_{\ell+1}\bigr),
\qquad
\frac{\partial E_{\ell+1}}{\partial a_{\ell}} = 0 .
\end{equation}
The first reads the target evidence of a predicted state from $\mathbf{h}^{O}_{\ell+1}$, the observation-branch tokens that generate the predicted view $\widehat{Z}_{\ell+1}$, with an evidence head distilled from the detector, $g_{\psi}(\mathbf{h}^{O}_{\ell+1}) \approx [H^{\mathrm{vis}}_{\ell+1}, (H^{\mathrm{area}}_{\ell+1})^{1/2}]$. Evidence thus follows the view, and, as with WoMAP's reward head~\citep{yin2025womap}, imagined states are scored without decoding. The second requires $E_{\ell+1}$, the transition's internal representation of the target, to be computed without the current action. Together they factorize the one-step transition as
\begin{equation}
\label{eq:latent_factor_main}
p(x_{\ell+1}, Z_{\ell+1}, H_{\ell+1} \mid S_{\ell}, a_{\ell})
= p(x_{\ell+1} \mid x_{\ell}, a_{\ell})\;
p(Z_{\ell+1} \mid Z_{\ell}, E_{\ell+1}, R_{\ell+1})\;
p(H_{\ell+1} \mid Z_{\ell+1}),
\end{equation}
with $E_{\ell+1} = f_E(H_{\ell}, Z_{\ell}, x_{\ell})$, $R_{\ell+1}$ the robot representation defined below, and $p(H_{\ell+1} \mid Z_{\ell+1})$ realized by $g_{\psi}$ on the tokens that generate $Z_{\ell+1}$. The constraint removes only the direct edge: the action still shapes all later target evidence through the view, so candidates that move the camera differently still receive different evidence. The mask fixes the route; Section~\ref{sec:offline} tests how the model ranks actions the behavior policy never took, under paired interventions from the same simulator state.

\subsubsection{Transition Architecture}
The transition realizes Eq.~\eqref{eq:latent_factor_main} as a stack of causal layers. Each layer updates three groups of tokens in a fixed order, and attention masks decide what each group may read. The target-evidence tokens are updated first, $E_{\ell+1} = f_E(H_{\ell}, Z_{\ell}, x_{\ell})$, and attend only to target-evidence, current-pose, and current-observation tokens, none of which carries $a_{\ell}$. The robot tokens are updated next, $x_{\ell+1} = f_x(x_{\ell}, a_{\ell})$ and $R_{\ell+1} = [\, x_{\ell+1}, W_a(a_{\ell})\,]$, so $R_{\ell+1}$ is the only representation that carries the current action. The observation tokens come last, $Z_{\ell+1} = f_O(Z_{\ell}; R_{\ell+1}, E_{\ell+1})$, where $f_E$ and $f_O$ are attention and feed-forward blocks. The transition $\mathcal{T}_{\theta} : (S_{\ell}, a_{\ell}) \mapsto S_{\ell+1}$ thus has the paths $a_{\ell} \rightarrow R_{\ell+1} \rightarrow Z_{\ell+1} \rightarrow H_{\ell+1}$ and $H_{\ell} \rightarrow E_{\ell+1} \rightarrow Z_{\ell+1}$ and none from $a_{\ell}$ to $E_{\ell+1}$, so the zero Jacobian of Eq.~\eqref{eq:readout} holds by construction (Section~\ref{sec:structural}).

\subsubsection{Training}
We train on transitions $(O_{\ell}, a_{\ell}, O_{\ell+1})$ from EVT-Bench~\citep{trackvla} and Habitat 3.0~\citep{h3} (Appendix~\ref{app:exp_setting}). Only the next observation latent is generated, since the next pose follows from kinematics and the next evidence from the head. We noise $Z_{\ell+1}$ under a DDPM schedule, $\tilde{Z}_u = \sqrt{\bar{\gamma}_u}\, Z_{\ell+1} + \sqrt{1-\bar{\gamma}_u}\, \varepsilon$ with $\varepsilon \sim \mathcal{N}(0, I)$, and train the denoiser $\epsilon_{\theta}$ with
\begin{equation}
\label{eq:diff_obj_main}
\mathcal{L}_{\mathrm{diff}}
=
\mathbb{E}_{\ell, u, \varepsilon}
\Bigl[
\lambda(u)\,
\bigl\|
\varepsilon - \epsilon_{\theta}(\tilde{Z}_u; S_{\ell}, a_{\ell})
\bigr\|_2^2
\Bigr].
\end{equation}
The evidence head is trained on the same forward pass. It average-pools the final-layer observation tokens $\mathbf{h}^{O}_{\ell+1}$ at noise level $u$ and regresses the detector's reading of the real next frame, $\mathcal{L}_{\mathrm{ev}} = \mathbb{E}_{\ell,u,\varepsilon}\, \operatorname{SL}_1\bigl(g_{\psi}(\mathbf{h}^{O}_{\ell+1}) - [H^{\mathrm{vis}}_{\ell+1}, (H^{\mathrm{area}}_{\ell+1})^{1/2}]\bigr)$ with $H_{\ell+1} = \Gamma(O_{\ell+1})$ and $\operatorname{SL}_1$ the smooth-$\ell_1$ loss; the square root spreads out the small boxes of distant targets. Because the observation branch attends to $E_{\ell+1}$ and $R_{\ell+1}$, this gradient reaches all three branches and shapes the dynamics toward what the planner reads. A small head $g_{\mathrm{dist}}$ on $E_{\ell+1}$, used only in training, predicts whether the next distance, taken from the simulator, lies in the following range (the auxiliary distance-aware loss $\mathcal{L}_{\mathrm{dist}}$, Appendix~\ref{app:method_details}). The total loss is $\mathcal{L} = \mathcal{L}_{\mathrm{diff}} + \lambda_{\mathrm{ev}} \mathcal{L}_{\mathrm{ev}} + \lambda_{\mathrm{dist}} \mathcal{L}_{\mathrm{dist}}$ with $\lambda_{\mathrm{ev}} = 1$ and $\lambda_{\mathrm{dist}} = 0.2$ (Appendix~\ref{app:method_details}). A rollout applies $\widehat{S}_{\ell+1} = \mathcal{T}_{\theta}(\widehat{S}_{\ell}, a_{\ell})$ recursively, and the head reads each step's evidence at $u=0$.

\subsection{Planning with the World Model}
\label{sec:planning}

Given the current state $S_0 = [H_0, Z_0, x_0]$, we search for an action sequence whose imagined future keeps the target in view at the right scale, or brings it back into view, without unsafe motion. We use the Cross-Entropy Method (CEM)~\citep{CEM} with horizon $T=10$, $N=128$ candidates, $K=16$ elites, and $M=4$ iterations, with $S=20$ DDIM sampling steps per imagined step. Each iteration samples $a_{0:T-1}^{(i)} \sim \mathcal{N}(\mu^{(m)}, \Sigma^{(m)})$, rolls out all candidates as one batch, scores them, and refits $(\mu^{(m+1)}, \Sigma^{(m+1)})$ to the elites (Appendix~\ref{app:method_details}). A candidate is scored by
\begin{equation}
\label{eq:reward}
\begin{aligned}
\mathcal{V}(S_0, a_{0:T-1})
={}& \sum_{t=1}^{T} \Bigl[
\beta_{\mathrm{vis}}\, \widehat{H}^{\mathrm{vis}}_{t}
- \beta_{\mathrm{dist}}\, \bigl|\widehat{H}^{\mathrm{area}}_{t} - \alpha\bigr|
\Bigr] \\
&- \lambda_{\mathrm{valid}} \sum_{t=0}^{T-1} \mathbb{I}\bigl(a_t \notin \mathcal{A}_{\mathrm{valid}}\bigr)
- \lambda_{\mathrm{safe}} \sum_{t=1}^{T} \mathbb{I}\bigl(\widehat{S}_{t} \notin \mathcal{O}_{\mathrm{safe}}\bigr),
\end{aligned}
\end{equation}
where $\widehat{H}_{t} = [\widehat{H}^{\mathrm{vis}}_{t}, \widehat{H}^{\mathrm{area}}_{t}] = g_{\psi}(\mathbf{h}^{O}_{t})$ is read by the evidence head from each imagined view, $\widehat{H}^{\mathrm{area}}_{t}$ being its estimate of $(H^{\mathrm{area}}_{t})^{1/2}$, and GroundingDINO runs only on the current observation $O_0$. The first term rewards seeing the target, the second rewards an apparent scale near the reference $\alpha$, and the penalties count actions outside the velocity and smoothness limits $\mathcal{A}_{\mathrm{valid}}$ and predicted states outside $\mathcal{O}_{\mathrm{safe}}$, a kinematic clearance proxy on the predicted robot displacement with margin $d_{\mathrm{safe}}=0.20$\,m that uses no map or depth (Appendix~\ref{app:method_details}). We set $\beta_{\mathrm{vis}}=\beta_{\mathrm{dist}}=\lambda_{\mathrm{valid}}=\lambda_{\mathrm{safe}}=1$ and $\alpha=0.5$. The robot executes the first action of the best sequence and replans at the next step (pseudocode in Appendix~\ref{app:algorithms}).

\paragraph{Ranking candidates while the target is hidden.}
When the target is out of sight, $H^{\mathrm{vis}}_{0} \approx 0$, yet the candidates still predict different views, because each moves the camera differently. The observation branch combines that motion with the target-evidence branch, which carries the last evidence and the context in which the target was seen, so a candidate whose imagined views show the target again at step $t$ collects the visibility and scale terms from $t$ on. With the shortcut, candidates would instead be ranked by their resemblance to the behavior policy. If no candidate brings the target back within $T$ steps, only the penalties remain, and the robot takes a safe first action and replans.

\section{Experiments}
\label{sec:experiments}

The experiments follow the chain of our claims: CST-WM tracks and recovers targets better than reactive trackers and an unstructured world model planned the same way (Sections~\ref{sec:main_results}--\ref{sec:recovery}); it does so because its imagined futures rank counterfactual candidates more like the real world (Section~\ref{sec:offline}), and they do so because the action is kept out of the target-evidence branch (Sections~\ref{sec:structural} and~\ref{sec:ablation}).

\subsection{Experimental Setting}
\label{sec:exp_setting}

\paragraph{Benchmarks and metrics.}
We use EVT-Bench~\citep{trackvla} and Habitat 3.0~\citep{h3} with egocentric observations and scene-disjoint splits. On EVT-Bench we evaluate single-target tracking (STT) with success rate (SR), tracking rate (TR), and collision rate (CR). On Habitat 3.0 we evaluate tracking from a visible target with following rate (F), distance-range success (DRS), CR, and episode success (ES), and recovery under simulated short and long occlusions, out-of-view drift, and distractor crossings with re-acquisition success (Re-acq.), time to re-acquire in steps (TTR), following rate after re-acquisition (Post-F), and recovery episode success (Rec-ES). Transfer trains on EVT-Bench and tests on Habitat 3.0. We report means over three seeds (Appendix~\ref{app:exp_setting}).

\begin{table*}[t]
\centering
\caption{Main results. Left: EVT-Bench, single-target tracking (STT). Middle: Habitat 3.0, standard tracking from a visible target. Right: transfer from EVT-Bench to Habitat 3.0. $^{\dagger}$Released pipeline without the diffusion planner (Appendix~\ref{app:exp_setting}). Best in \textbf{bold}, second \underline{underlined}.}
\label{tab:main_quantitative}
\setlength{\tabcolsep}{4.8pt}
\renewcommand{\arraystretch}{1.08}
\resizebox{\textwidth}{!}{%
\begin{tabular}{lccc lcccc lcccc}
\toprule
\multicolumn{4}{c}{EVT-Bench (STT)} &
\multicolumn{5}{c}{Habitat 3.0, standard tracking} &
\multicolumn{5}{c}{Cross-dataset transfer to Habitat 3.0} \\
\cmidrule(lr){1-4}\cmidrule(lr){5-9}\cmidrule(lr){10-14}
Method & SR$\uparrow$ & TR$\uparrow$ & CR$\downarrow$ &
Method & F$\uparrow$ & DRS$\uparrow$ & CR$\downarrow$ & ES$\uparrow$ &
Method & F$\uparrow$ & DRS$\uparrow$ & CR$\downarrow$ & ES$\uparrow$ \\
\midrule
Uni-NaVid & 25.7 & 39.5 & 41.9 &
Habitat 3.0 baseline & 0.29 & 0.47 & 0.48 & 0.40 &
Uni-NaVid & 0.40 & 0.56 & 0.39 & 0.45 \\
OA-VAT$^{\dagger}$ & 32.4 & 63.9 & 3.70 &
Uni-NaVid & 0.34 & 0.52 & 0.43 & 0.47 &
TrackVLA & 0.38 & 0.54 & 0.41 & 0.43 \\
TrackVLA & 85.1 & 78.6 & \underline{1.65} &
Adapted NWM & \underline{0.41} & 0.61 & \underline{0.33} & \underline{0.49} &
Adapted NWM & \underline{0.43} & 0.58 & \underline{0.35} & \underline{0.47} \\
TrackVLA++ & \underline{86.0} & \underline{81.0} & 2.10 &
SDA-S2 & 0.39 & \underline{0.63} & 0.57 & 0.43 &
TrackVLA++ & 0.42 & \underline{0.61} & 0.38 & 0.43 \\
Ours & \textbf{88.7} & \textbf{83.4} & \textbf{1.41} &
Ours & \textbf{0.53} & \textbf{0.70} & \textbf{0.27} & \textbf{0.61} &
Ours & \textbf{0.48} & \textbf{0.65} & \textbf{0.30} & \textbf{0.54} \\
\bottomrule
\end{tabular}%
}
\end{table*}

\paragraph{Baselines.}
The first group consists of complete tracking systems with their own inputs, training data, and action selection: Uni-NaVid~\citep{uninavid}, TrackVLA~\citep{trackvla}, TrackVLA++~\citep{trackvlapp}, the Habitat 3.0 baseline~\citep{h3}, SDA-S2~\citep{sda}, and OA-VAT~\citep{sun2026oavat}, an occlusion-aware active tracker that we run on EVT-Bench STT with its released pipeline and PID control. The second group isolates our contribution. The adapted NWM~\citep{nwm} keeps NWM's non-factorized CDiT-XL/2 transition, in which the action conditions every token, and shares data, optimizer, evidence read-out, objective (Eq.~\eqref{eq:reward}), CEM budget, and action bounds with CST-WM, at 6.6$\times$ our parameters and 3.6$\times$ our compute per pass, so any difference comes from the transition. Table~\ref{tab:baseline_alignment} in Appendix~\ref{app:exp_setting} lists what every method receives.

\subsection{Quantitative Comparison}
\label{sec:main_results}

Table~\ref{tab:main_quantitative} reports the three settings. On EVT-Bench STT, CST-WM reaches 88.7 SR and 83.4 TR, 2.7 and 2.4 points above TrackVLA++, with the lowest collision rate (1.41). On Habitat 3.0 it leads on all four metrics (F 0.53, DRS 0.70, CR 0.27, ES 0.61, against best baselines of 0.41, 0.63, 0.33, and 0.49). Against the adapted NWM, which differs only in the transition, the factorized model improves F and ES by 0.12 each and lowers CR by 0.06, although the adapted NWM is the larger model. The advantage carries over to transfer (+0.05 F, +0.07 DRS, +0.07 ES, $-$0.05 CR).

\subsection{Target Loss and Re-acquisition}
\label{sec:recovery}

\begin{figure}[t]
  \centering
  \begin{minipage}[b]{0.485\linewidth}
    \centering
    \includegraphics[width=\linewidth]{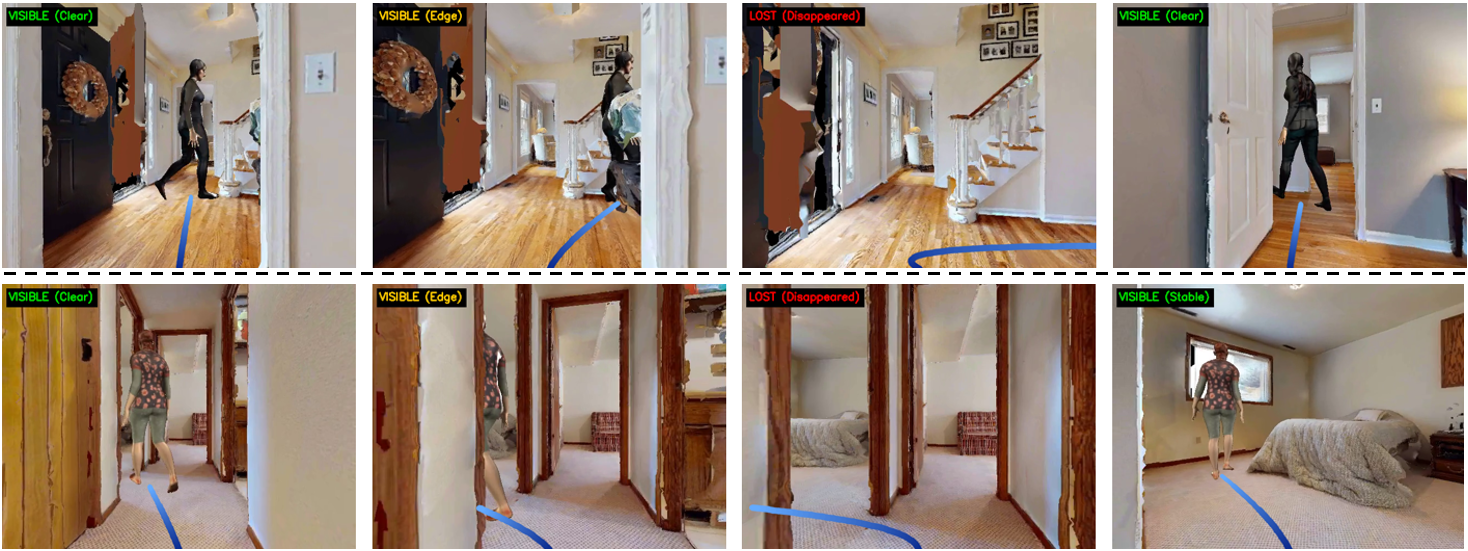}
    \captionof{figure}{Recovery with CST-WM (Habitat 3.0 target-loss protocol). Each row is one episode: visible, at the image edge, lost, re-acquired; the blue curve is the planned path.}
    \label{fig:simulate_track}
  \end{minipage}\hfill
  \begin{minipage}[b]{0.485\linewidth}
    \centering
    \includegraphics[width=\linewidth]{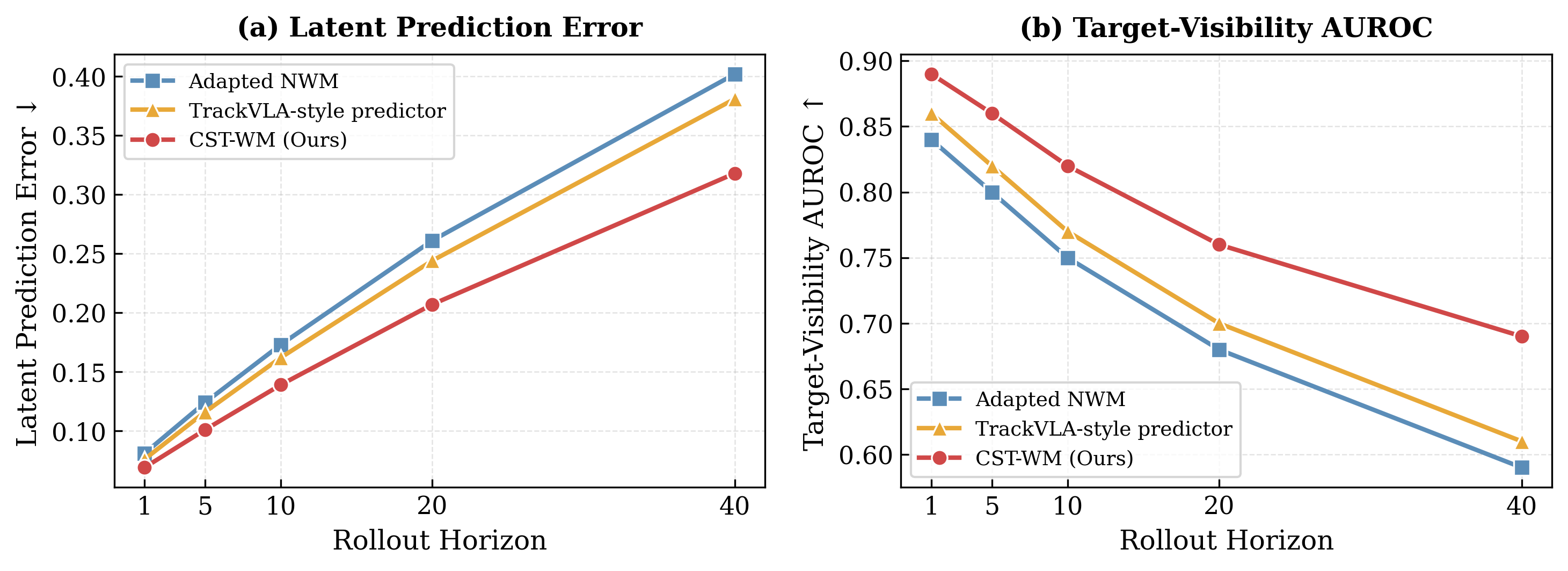}
    \captionof{figure}{Offline rollout fidelity against horizon: (a)~latent prediction error, (b)~target-visibility AUROC.}
    \label{fig:offline_rollout_vis}
  \end{minipage}
\end{figure}

\begin{table*}[t]
\centering
\caption{Recovery after temporary target loss (Habitat 3.0 target-loss protocol). Re-acq.: re-acquisition success; TTR: time to re-acquire (steps); Post-F: following rate after re-acquisition; Rec-ES: recovery episode success. Best in \textbf{bold}, second \underline{underlined}.}
\label{tab:reacquisition}
\setlength{\tabcolsep}{4.6pt}
\renewcommand{\arraystretch}{1.05}
\resizebox{\textwidth}{!}{%
\begin{tabular}{lcccccccc}
\toprule
& \multicolumn{2}{c}{Short occlusion} & \multicolumn{2}{c}{Long occlusion} & \multicolumn{2}{c}{Out-of-view drift} & \multicolumn{2}{c}{Distractor crossing} \\
\cmidrule(lr){2-3}\cmidrule(lr){4-5}\cmidrule(lr){6-7}\cmidrule(lr){8-9}
Method & Re-acq.$\uparrow$ & TTR$\downarrow$ & Re-acq.$\uparrow$ & TTR$\downarrow$ & Re-acq.$\uparrow$ & TTR$\downarrow$ & Re-acq.$\uparrow$ & TTR$\downarrow$ \\
\midrule
TrackVLA & 0.71 & 8.9 & 0.48 & 14.6 & 0.52 & 12.8 & 0.43 & 15.2 \\
Adapted NWM & \underline{0.75} & \underline{8.1} & \underline{0.54} & \underline{13.2} & \underline{0.57} & \underline{11.7} & \underline{0.49} & \underline{14.0} \\
Ours & \textbf{0.84} & \textbf{6.4} & \textbf{0.69} & \textbf{9.8} & \textbf{0.73} & \textbf{8.7} & \textbf{0.65} & \textbf{10.6} \\
\midrule
Method & Post-F$\uparrow$ & Rec-ES$\uparrow$ & Post-F$\uparrow$ & Rec-ES$\uparrow$ & Post-F$\uparrow$ & Rec-ES$\uparrow$ & Post-F$\uparrow$ & Rec-ES$\uparrow$ \\
\midrule
TrackVLA & 0.58 & 0.47 & 0.41 & 0.29 & 0.45 & 0.33 & 0.37 & 0.24 \\
Adapted NWM & \underline{0.62} & \underline{0.50} & \underline{0.45} & \underline{0.34} & \underline{0.49} & \underline{0.37} & \underline{0.41} & \underline{0.29} \\
Ours & \textbf{0.71} & \textbf{0.61} & \textbf{0.57} & \textbf{0.47} & \textbf{0.60} & \textbf{0.50} & \textbf{0.54} & \textbf{0.43} \\
\bottomrule
\end{tabular}%
}
\end{table*}

For each cause of target loss (Table~\ref{tab:reacquisition}), CST-WM has the highest Re-acq., Post-F, and Rec-ES and the shortest TTR under all four conditions. Its margin over the adapted NWM is 0.09 in Re-acq.\ under short occlusion and 0.15--0.16 under long occlusion, out-of-view drift, and distractor crossing, where the target stays hidden longest; it also re-acquires in 6.4--10.6 steps against 8.1--14.0. With objective and budget shared, the gap lies in which candidates each model imagines bringing the target back (Figure~\ref{fig:simulate_track}).

\subsection{Offline World-Model Evaluation}
\label{sec:offline}

Causal hallucination shows up only under counterfactual actions (Section~\ref{sec:causal-diffusion}), so we compare models offline under paired interventions: from each of 500 saved simulator states, the same 32 candidate action sequences are executed in the simulator and imagined by every model, including a TrackVLA-style predictor (a non-factorized transition conditioned on the action at every token); each model reads imagined evidence with its own head.

\subsubsection{Multi-Step Rollout Fidelity}
We compare predicted and simulated futures at horizons of 1, 5, 10, 20, and 40 steps. CST-WM has lower latent prediction error and higher target-visibility AUROC, scored against GroundingDINO on the simulated frames, at every horizon (Figure~\ref{fig:offline_rollout_vis}). The gap grows with the horizon, so the structured transition accumulates less error over the multi-step rollouts that planning relies on.

\begin{table}[t]
\centering
\begin{minipage}[t]{0.43\linewidth}
\centering
\caption{Offline diagnostics (Habitat 3.0). Top: ranking vs.\ simulator, the Plan.\,$\rho$ of Tables~\ref{tab:extra_ablation_breakdown}--\ref{tab:readout_closedloop} (TrackVLA-style: $\rho$ 0.52, $\tau$ 0.37). Middle: action in $E_{\ell+1}$. Bottom: humanoid still, robot moving.}
\label{tab:structural}
\setlength{\tabcolsep}{4pt}
\footnotesize
\resizebox{\linewidth}{!}{%
\begin{tabular}{lcc}
\toprule
Diagnostic & Adapted NWM & Ours \\
\midrule
Ranking: Spearman $\rho$$\uparrow$ & 0.47 & \textbf{0.68} \\
Ranking: Kendall $\tau$$\uparrow$ & 0.33 & \textbf{0.51} \\
\midrule
$|\partial E_{\ell+1}/\partial a_{\ell}|$$\downarrow$ & 0.112 & \textbf{0.001} \\
Var.\ under action swap$\downarrow$ & 0.084 & \textbf{0.006} \\
$W_1$ under action swap$\downarrow$ & 0.127 & \textbf{0.011} \\
\midrule
Stability of $E_{\ell+1}$$\uparrow$ & 0.61 & \textbf{0.98} \\
View sensitivity$\uparrow$ & 0.78 & \textbf{0.84} \\
\bottomrule
\end{tabular}%
}
\end{minipage}\hfill
\begin{minipage}[t]{0.54\linewidth}
\centering
\caption{Reading imagined target evidence on held-out 8-step rollouts, against GroundingDINO on the real frame: visibility AUROC after 1 and 8 steps, confidence MAE after 1 step, and $\Delta_{\mathrm{turn}}$, the change of predicted confidence after 8 steps when every turn is reversed (not ranked).}
\label{tab:evidence_readout}
\setlength{\tabcolsep}{2pt}
\footnotesize
\begin{tabular}{lcccc}
\toprule
Read-out & 1 step$\uparrow$ & 8 steps$\uparrow$ & MAE$\downarrow$ & $\Delta_{\mathrm{turn}}$ \\
\midrule
Copy current evidence & \textbf{0.936} & 0.589 & \textbf{0.054} & -- \\
Detect on decoded view & 0.730 & 0.512 & 0.276 & -- \\
Separate latent head & 0.875 & 0.623 & \underline{0.113} & -- \\
Head, first sampling step & \underline{0.881} & \underline{0.693} & 0.128 & 0.030 \\
Head, $u{=}0$ (ours) & 0.880 & 0.675 & 0.130 & 0.034 \\
Head, w/o masking & 0.860 & \textbf{0.701} & 0.131 & 0.025 \\
\bottomrule
\end{tabular}
\end{minipage}
\end{table}

\subsubsection{Planning-Value Consistency}
Accurate futures help only if they order candidates correctly. For each state we score its 32 candidates with Eq.~\eqref{eq:reward} and compare the ranking with the one given by their outcomes in the simulator from that state. CST-WM agrees best with the simulator (Spearman $\rho$ 0.68 against 0.47 for the adapted NWM and 0.52 for the TrackVLA-style predictor; Table~\ref{tab:structural}).

\subsubsection{Reading Imagined Target Evidence}
The ranking rests on what the evidence head reads from imagined states. On held-out 8-step rollouts under logged actions, we compare each read-out with GroundingDINO on the real future frame (Table~\ref{tab:evidence_readout}; protocol in Appendix~\ref{app:evidence_readout}). After one step, copying the current evidence is hard to beat, since the view barely changes; after 8 steps the evidence head reaches a visibility AUROC of 0.675, against 0.589 for copying, 0.512 for detecting on the decoded view, and 0.623 for a separately trained latent head. Reversing the turn of every action changes the head's predicted confidence after 8 steps by 0.034 with the mask and by 0.025 without it; with the mask it can only arise through the view. Without the mask the head predicts the logged future better after 8 steps (0.701 against 0.675), yet it ranks the counterfactual candidates worse ($\rho$ 0.55 against 0.68, Table~\ref{tab:extra_ablation_breakdown}) and recovers worse in closed loop (0.61 against 0.76). Accuracy on the actions the behavior policy took does not carry over to the actions it did not, which is the pattern a shortcut predicts.

\subsection{Structural and Invariance Analysis}
\label{sec:structural}

The offline gains should come from the constraint of Section~\ref{sec:causal-diffusion}, not from a better backbone. Two diagnostics check it: the Jacobian $|\partial E_{\ell+1}/\partial a_{\ell}|$ and the change of $E_{\ell+1}$ when $a_{\ell}$ is swapped for other valid actions (Table~\ref{tab:structural}). Both are near zero for CST-WM, as the mask requires, whereas the adapted NWM and leaky variants absorb action information when allowed to (mean Jacobian 0.11 and up to 0.33). The mask does not guarantee a third property: with the humanoid held still and the robot moving, $E_{\ell+1}$ can change through $Z_{\ell}$ and $x_{\ell}$. CST-WM keeps it stable (0.98 against 0.61) while its predicted views follow the viewpoint (0.84 against 0.78). The action thus moves the view but not the target.

\subsection{Runtime and Real-World Tracking}
\label{sec:runtime}

\begin{figure}[t]
  \centering
  \includegraphics[width=\linewidth]{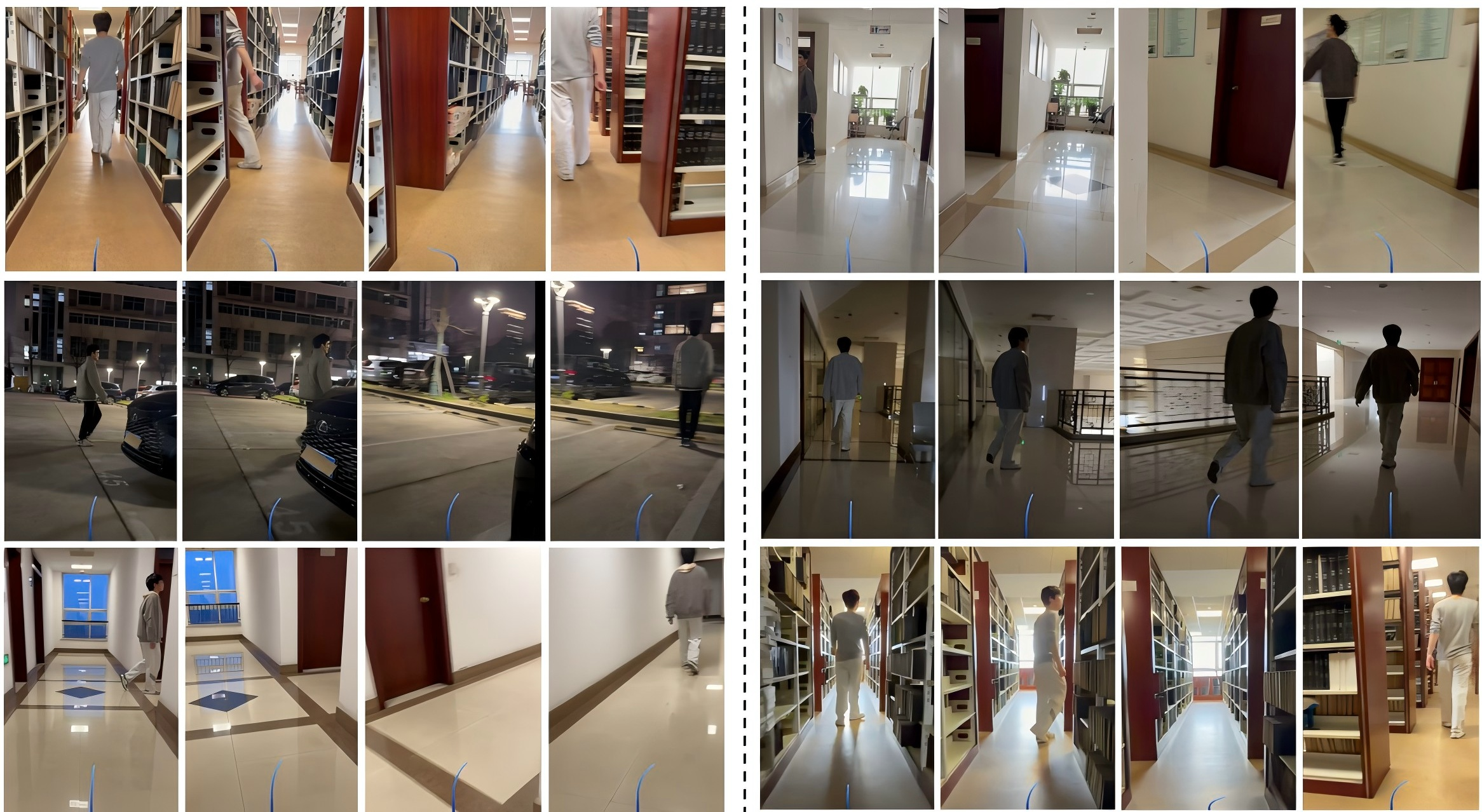}
  \caption{Real-world tracking with CST-WM in a library, office corridors, and a parking lot at night. Each row is one episode, left to right in time; the blue curve is the planned path.}
  \label{fig:realworld}
\end{figure}

\begin{figure}[t]
  \centering
  \begin{minipage}[c]{0.46\linewidth}
    \centering
    \includegraphics[width=\linewidth]{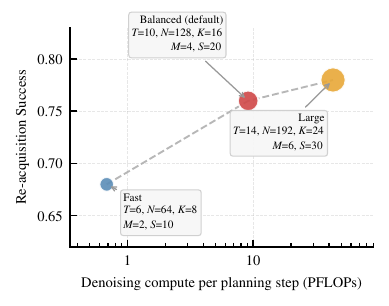}
    \captionof{figure}{Re-acquisition against denoising compute per planning step; bubble size $\propto N$.}
    \label{fig:runtime_pareto}
  \end{minipage}\hfill
  \begin{minipage}[c]{0.5\linewidth}
    \centering
    \captionof{table}{Real-world tracking, 10 trials per scenario and method. SR/TR: success/tracking rate (\%); 95\% intervals in Appendix~\ref{app:exp_setting}.}
\label{tab:realworld}
\setlength{\tabcolsep}{4pt}
\footnotesize
\begin{tabular}{lcccc}
\toprule
& \multicolumn{2}{c}{TrackVLA} & \multicolumn{2}{c}{Ours} \\
\cmidrule(lr){2-3}\cmidrule(lr){4-5}
Scenario & SR$\uparrow$ & TR$\uparrow$ & SR$\uparrow$ & TR$\uparrow$ \\
\midrule
Occlusion & 40 & 55.8 & \textbf{80} & \textbf{64.1} \\
Distractor & \textbf{60} & 47.9 & \textbf{60} & \textbf{56.2} \\
Fast motion & 40 & 43.5 & \textbf{60} & \textbf{51.2} \\
\bottomrule
\end{tabular}
  \end{minipage}
\end{figure}

\paragraph{Planning cost.}
A planning step runs $MTS$ batched denoising passes over the $N$ candidates, so its wall-clock time follows the budget. On one RTX A6000 a pass over 128 candidates takes 0.36\,s, 2.8$\times$ faster than NWM's backbone. From observation to action, the default budget of 800 passes takes 267\,s, the fast budget of Figure~\ref{fig:runtime_pareto} 22.5\,s, and the robot budget below 8.2\,s (Table~\ref{tab:runtime_e2e}). The evidence head reads 128 imagined views in 0.2\,ms, against 11\,ms and 38\,ms per view to decode and detect. The simulator waits for the planner, so these times do not enter the simulated results, and a 4.7$\times$ larger budget adds only 0.02 re-acquisition. Under NWM's single-trajectory protocol, CST-WM matches NWM's reported times on the same GPU (Appendix~\ref{app:runtime}).

\paragraph{Real-world tracking.}
We follow the real-robot protocol of TrackVLA~\citep{trackvla}: occlusion, distractor crossing, and fast motion, with 10 trials per scenario and method (Appendix~\ref{app:exp_setting}). The robot is a Unitree Go2 with an Intel RealSense D435i RGB camera; as in TrackVLA and OA-VAT, planning runs on an offboard RTX 6000 Ada GPU with a reduced budget ($T{=}4$, $N{=}24$, $K{=}8$, $M{=}2$, 10 sampling steps), one planning step every 1.4\,s; for deployment, the evidence head is further trained on 100k real frames of people from public videos together with 100k simulator frames (Appendix~\ref{app:exp_setting}). The model keeps no target identity, so a trial in which the robot follows the distractor counts as a failure. CST-WM succeeds in 8, 6, and 6 of 10 trials, against 4, 6, and 4 for TrackVLA (Table~\ref{tab:realworld}; episodes in Figure~\ref{fig:realworld} and the video).

\subsection{Ablation Study}
\label{sec:ablation}

\begin{table}[t]
\centering
\begin{minipage}[t]{0.51\linewidth}
\centering
\caption{One-component ablations, Habitat 3.0.}
\label{tab:extra_ablation_breakdown}
\setlength{\tabcolsep}{2.4pt}
\footnotesize
\begin{tabular}{lcccc}
\toprule
Variant & F$\uparrow$ & DRS$\uparrow$ & Re-acq.$\uparrow$ & Plan.\,$\rho\uparrow$ \\
\midrule
W/o masking & 0.47 & 0.63 & 0.61 & 0.55 \\
W/o factorization & 0.49 & 0.66 & 0.64 & 0.58 \\
W/o dist.-aware loss & 0.49 & 0.65 & \underline{0.72} & \underline{0.64} \\
Deterministic & \underline{0.50} & \underline{0.67} & 0.69 & 0.60 \\
Full model & \textbf{0.53} & \textbf{0.70} & \textbf{0.76} & \textbf{0.68} \\
\bottomrule
\end{tabular}
\end{minipage}\hfill
\begin{minipage}[t]{0.47\linewidth}
\centering
\caption{Architectural variants, Habitat 3.0.}
\label{tab:arch_ablation}
\setlength{\tabcolsep}{2.2pt}
\footnotesize
\begin{tabular}{lcccc}
\toprule
Variant & F$\uparrow$ & Re-acq.$\uparrow$ & Plan.\,$\rho\uparrow$ & CR$\downarrow$ \\
\midrule
Partial masking & 0.46 & 0.68 & \underline{0.62} & \underline{0.30} \\
Shuffled order & 0.46 & 0.65 & 0.58 & 0.32 \\
Deterministic & \underline{0.50} & \underline{0.69} & 0.60 & \underline{0.30} \\
Full model & \textbf{0.53} & \textbf{0.76} & \textbf{0.68} & \textbf{0.27} \\
\bottomrule
\end{tabular}
\end{minipage}
\end{table}

Each ablation changes one component (Table~\ref{tab:extra_ablation_breakdown}); w/o masking removes only the target-evidence mask. Removing the mask costs the most re-acquisition (0.76 to 0.61) and planning-value consistency (0.68 to 0.55). A single-branch transition lowers F and consistency, dropping the distance-aware loss lowers all four by 0.04--0.05, and a deterministic predictor weakens recovery. Partial masking and a shuffled order also lower re-acquisition (Table~\ref{tab:arch_ablation}); seeds vary by $\le$0.02 (Table~\ref{tab:seed_breakdown}).

\paragraph{Target evidence: source and read-out.}
Table~\ref{tab:readout_closedloop} varies where the planner's target evidence comes from. Our signal comes close to the simulator's true distance (0.76 against 0.78 re-acquisition). Detecting on simulator-rendered future views gives the highest planning-value consistency (0.74) and the lowest CR (0.24), but, reading raw detector scores, it reaches F and re-acquisition (0.48, 0.64) close to the raw-score row rather than our trained head. Among the deployable read-outs, the jointly trained head, whose loss also shapes the dynamics as in WoMAP~\citep{yin2025womap}, leads on every metric (re-acquisition 0.76, against 0.43 on a frozen transition, 0.57 for a separate latent head, and 0.46 for decode-and-detect) and reads an imagined step in 0.018\,s instead of 5.8\,s. The raw detector score lowers DRS to 0.60 and re-acquisition to 0.66; the distance-aware loss raises the signal's rank correlation with true distance from 0.71 to 0.83 across humanoids (Table~\ref{tab:cross_humanoid}).

\begin{table}[!h]
\centering
\caption{Target evidence for planning on Habitat 3.0: privileged sources, imagined-step read-outs, and signal variants; transition, objective, and CEM budget fixed. Time: read-out per imagined step.}
\label{tab:readout_closedloop}
\setlength{\tabcolsep}{3.6pt}
\footnotesize
\begin{tabular}{llcccccc}
\toprule
& Target evidence & F$\uparrow$ & DRS$\uparrow$ & Re-acq.$\uparrow$ & CR$\downarrow$ & Plan.\,$\rho\uparrow$ & Time (s)$\downarrow$ \\
\midrule
\multirow{2}{*}{Privileged} & Simulator true distance & 0.54 & 0.72 & 0.78 & 0.26 & -- & -- \\
 & Detector on rendered future views & 0.48 & 0.67 & 0.64 & 0.24 & 0.74 & 5.2 \\
\midrule
\multirow{3}{*}{Read-out} & Detector on decoded views & 0.38 & 0.46 & 0.46 & 0.40 & 0.58 & 5.8 \\
 & Separate latent head on $\widehat{Z}$ & 0.45 & 0.60 & 0.57 & 0.33 & 0.54 & 0.030 \\
 & Head, frozen transition & 0.34 & 0.37 & 0.43 & 0.39 & 0.62 & 0.021 \\
\midrule
\multirow{2}{*}{Signal} & Raw detector score & 0.46 & 0.60 & 0.66 & 0.32 & -- & -- \\
 & W/o distance-aware loss & 0.49 & 0.65 & 0.72 & 0.29 & 0.64 & -- \\
\midrule
Ours & Evidence head, joint training & 0.53 & 0.70 & 0.76 & 0.27 & 0.68 & 0.018 \\
\bottomrule
\end{tabular}
\end{table}

\FloatBarrier
\section{Conclusion}
\label{conclusion}

CST-WM plans embodied visual tracking over imagined target evidence: its transition keeps the current action out of the target-evidence branch, so the action enters only the robot branch and reaches later evidence through the views it produces, and a detector-distilled head reads the evidence of each imagined future from the predicted view. Under paired interventions from the same simulator state, its imagined futures rank candidates closer to their simulated outcomes than the transition without the mask ($\rho$ 0.68 against 0.55), and the gain carries to closed loop (re-acquisition 0.76 against 0.61). One model improves following, safety, and re-acquisition on EVT-Bench and Habitat 3.0, transfers from the first to the second, and holds on a real robot.

\paragraph{Limitations.}
CST-WM plans over detector-derived evidence, not the target's state: it inherits detector errors, keeps no target identity, and does not track a hidden target past the horizon (Appendix~\ref{app:limitations}). Its transition conditions on the current view only, so the target's motion is inferred from one frame, and each planning step runs its denoising passes in sequence, so the robot budget trades horizon and candidates for time (Appendix~\ref{app:runtime}).


\subsection*{Reproducibility statement}
The structured state, the causal transition, the evidence head, and the planning objective are specified in Section~\ref{sec:causal-diffusion} and Section~\ref{sec:planning}, with the full loss and CEM update equations in Appendix~\ref{app:method_details} and pseudocode in Appendix~\ref{app:algorithms}. Data splits, recovery protocols, baselines, and evaluation metrics are described in Appendix~\ref{app:exp_setting}, and additional results are reported in Appendix~\ref{app:more_results}.

\bibliographystyle{iclr2027_conference}
\bibliography{main}

\appendix
\clearpage
\setlength{\parskip}{0pt plus 1pt}
\setlength{\textfloatsep}{20pt plus 2pt minus 4pt}
\setlength{\floatsep}{12pt plus 2pt minus 2pt}
\setlength{\intextsep}{12pt plus 2pt minus 2pt}
\setlength{\dbltextfloatsep}{20pt plus 2pt minus 4pt}
\setlength{\dblfloatsep}{12pt plus 2pt minus 2pt}
\setlength{\abovecaptionskip}{10pt}
\setlength{\belowcaptionskip}{0pt}
\captionsetup{skip=10pt}
\setlength{\abovedisplayskip}{12pt plus 3pt minus 7pt}
\setlength{\belowdisplayskip}{12pt plus 3pt minus 7pt}
\setlength{\abovedisplayshortskip}{0pt plus 3pt}
\setlength{\belowdisplayshortskip}{7pt plus 3pt minus 4pt}
\titlespacing*{\section}{0pt}{3.5ex plus 1ex minus .2ex}{2.3ex plus .2ex}
\titlespacing*{\subsection}{0pt}{3.25ex plus 1ex minus .2ex}{1.5ex plus .2ex}
\titlespacing*{\subsubsection}{0pt}{3.25ex plus 1ex minus .2ex}{1.5ex plus .2ex}
\titlespacing*{\paragraph}{0pt}{3.25ex plus 1ex minus .2ex}{1em}

\appendix

\section{Method Details}
\label{app:method_details}

\paragraph{Auxiliary distance-aware loss.}
The full form of the auxiliary distance-aware loss attached to the target-evidence branch is
\begin{equation}
\label{eq:distance_loss_app}
\mathcal{L}_{\mathrm{dist}}
=
- \mathbb{E}_{\ell}\!\bigl[\,
y^{\mathrm{dist}}_{\ell+1}
\log g_{\mathrm{dist}}(E_{\ell+1})
+ (1-y^{\mathrm{dist}}_{\ell+1})
\log(1-g_{\mathrm{dist}}(E_{\ell+1}))
\,\bigr],
\end{equation}
where $y^{\mathrm{dist}}_{\ell+1} = \mathbb{I}(d^{-} \le d_{\ell+1} \le d^{+})$ is the binary in-range distance label derived from the simulator-only relative distance $d_{\ell+1}$, with $[d^{-}, d^{+}]$ the following range, and $g_{\mathrm{dist}}$ is a lightweight prediction head on the target-evidence branch $E_{\ell+1}$. The distance is used only in training; the planner does not use the head.

\paragraph{Evidence head.}
The labels of $\mathcal{L}_{\mathrm{ev}}$ are GroundingDINO (Swin-T) readings of the $224{\times}224$ training frames with the prompt ``humanoid'' and box and text thresholds of 0.20; the top-scoring box gives $H^{\mathrm{vis}}$ (its confidence) and $H^{\mathrm{area}}$ (its area over the image area). The head $g_{\psi}$ is LayerNorm, a $512{\times}512$ linear layer, GELU and a $512{\times}2$ linear layer with a sigmoid, applied to the mean of the $28{\times}28$ final-layer observation tokens (0.26M parameters). We add the head and $\mathcal{L}_{\mathrm{ev}}$ to a transition trained with $\mathcal{L}_{\mathrm{diff}}$ for 12k steps and continue training all parameters with the total loss of Section~\ref{sec:causal-diffusion} (smooth-$\ell_1$ with threshold 0.05 for $\mathcal{L}_{\mathrm{ev}}$). At planning time the head is read once per imagined step, from the tokens of the final denoising pass ($u = 0$), so no extra pass is run.

\paragraph{CEM update.}
At iteration $m$, with $\mathcal{I}^{(m)}$ denoting the indices of the top-$K$ candidates, the sampling distribution is updated by
\begin{equation}
\mu^{(m+1)}
=
\tfrac{1}{K}\!\sum_{i \in \mathcal{I}^{(m)}}\!\! a_{0:T-1}^{(i)},\quad
\Sigma^{(m+1)}
=
\tfrac{1}{K}\!\sum_{i \in \mathcal{I}^{(m)}}\!\!
\bigl(a_{0:T-1}^{(i)} - \mu^{(m+1)}\bigr)
\bigl(a_{0:T-1}^{(i)} - \mu^{(m+1)}\bigr)^{\!\top}.
\end{equation}

\paragraph{Admissible action and safe rollout sets.}
The admissible action set enforces platform velocity and smoothness limits,
\begin{equation}
\mathcal{A}_{\mathrm{valid}}
=
\bigl\{\, a_t = [v_t, \omega_t] \,\bigm|\,
\|v_t\|_{\infty} \le v_{\max},\;
|\omega_t| \le \omega_{\max},\;
\|a_t-a_{t-1}\|_{\infty} \le \delta_{\max}
\,\bigr\},
\end{equation}
with $a_{-1}$ the last executed action, and the safe rollout set is defined on the predicted robot branch alone, without map, depth, or simulator geometry. With $\widehat{p}_t$ the planar position in $\widehat{x}_t$ and $\Delta p_t = \|\widehat{p}_t - \widehat{p}_{t-1}\|_2$ the displacement predicted for one control period $\Delta t$, the clearance proxy is $d_{\mathrm{clr}}(\widehat{S}_t) = v_{\max}\Delta t + d_{\mathrm{safe}} - \Delta p_t$ and $\mathcal{O}_{\mathrm{safe}} = \{\, \widehat{S}_t \mid d_{\mathrm{clr}}(\widehat{S}_t) \ge 0 \,\}$ with $d_{\mathrm{safe}}=0.20$\,m: a predicted state is unsafe when its predicted displacement in one period exceeds the platform maximum $v_{\max}\Delta t$ by more than the margin $d_{\mathrm{safe}}$. The proxy is not a distance to obstacles. Motion limits are matched to the simulator bounds used by all baselines.

\FloatBarrier
\section{Algorithm Summaries}
\label{app:algorithms}

Algorithm~\ref{alg:training_cctwm} summarizes training and Algorithm~\ref{alg:planning_cctwm} planning.

\begin{table}[!ht]
\centering
\begin{minipage}[t]{0.49\linewidth}
\centering
\captionof{algorithm}{Training procedure for CST-WM.}
\label{alg:training_cctwm}
\setlength{\tabcolsep}{4pt}
\footnotesize
\begin{tabular}{p{0.94\linewidth}}
\toprule
Input: transition dataset $(O_{\ell}, a_{\ell}, O_{\ell+1})$, model parameters $\theta$, diffusion scheduler, optimizer. \\
Output: trained world model $\theta^{\star}$. \\
\midrule
1. Encode $O_{\ell}$ and $O_{\ell+1}$ with the visual encoder to obtain the observation latents $Z_{\ell}$ and $Z_{\ell+1}$. \\
2. Read the target evidence $H_{\ell}$ and $H_{\ell+1}$ from $O_{\ell}$ and $O_{\ell+1}$ with the detector (confidence and normalized target area). \\
3. Form the structured state $S_{\ell}=[H_{\ell}, Z_{\ell}, x_{\ell}]$, with $H_{\ell}$ in its scalar form, and the robot inputs $x_{\ell}$, $a_{\ell}$. \\
4. Sample a diffusion step and corrupt the next observation latent $Z_{\ell+1}$ with Gaussian noise. \\
5. Predict the diffusion noise with the masked transition network (target-evidence branch without $a_{\ell}$, robot branch with $a_{\ell}$, observation branch reading both). \\
6. Compute the denoising objective, the evidence-head objective $\mathcal{L}_{\mathrm{ev}}$ against $H_{\ell+1}$ on the pooled observation tokens, and the auxiliary distance-aware objective on the target-evidence branch (disabled only in the w/o distance-aware loss ablation); the head and $\mathcal{L}_{\mathrm{ev}}$ are added after a 12k-step warm start with the denoising objective alone. \\
7. Update model parameters with AdamW. \\
8. Update the exponential moving average of the parameters. \\
9. Repeat until convergence and keep the validation-best checkpoint. \\
\bottomrule
\end{tabular}
\end{minipage}\hfill
\begin{minipage}[t]{0.49\linewidth}
\centering
\captionof{algorithm}{MPC planning procedure for CST-WM with the Cross-Entropy Method.}
\label{alg:planning_cctwm}
\setlength{\tabcolsep}{4pt}
\footnotesize
\begin{tabular}{p{0.94\linewidth}}
\toprule
Input: current observation $O_0$ with its detector reading $H_0$, current state $S_0 = [H_0, Z_0, x_0]$, planning hyperparameters $T$, $N$, $K$, $M$, $S$. \\
Output: next control action $a_0^{\star}$. \\
\midrule
1. Initialize the Gaussian sampling distribution over action sequences. \\
2. For each CEM iteration, sample $N$ candidate action sequences. \\
3. Roll out the world model over horizon $T$ for all candidates as one batch; read the target evidence of every predicted step with the evidence head $g_{\psi}$. \\
4. Score each candidate with the planning objective (Eq.~\eqref{eq:reward}) on the evidence read from its imagined views. \\
5. Select the top-$K$ elite sequences and update the sampling distribution. \\
6. After the final iteration, choose the best sequence and execute its first action. \\
7. Replan in receding-horizon form at the next step. \\
\bottomrule
\end{tabular}
\end{minipage}
\end{table}

\section{Detailed Experimental Setting}
\label{app:exp_setting}

\paragraph{Benchmarks and tasks.}
We evaluate on EVT-Bench~\citep{trackvla} and Habitat 3.0~\citep{h3}. In both, the robot receives only egocentric observations and platform actions at test time; no humanoid state, future trajectory, or distance is available. We evaluate EVT-Bench's single-target tracking task (STT). Habitat 3.0 has richer scene geometry and more frequent occlusion and reappearance. On Habitat 3.0 we evaluate standard tracking from a visible target and a target-loss protocol in which the target becomes unavailable through short occlusion, long occlusion, out-of-view drift, or distractor crossing. For cross-dataset transfer, models trained on EVT-Bench are evaluated on Habitat 3.0 under the standard-tracking protocol.

\paragraph{Data construction and splits.}
For each benchmark we build one-step transitions $(O_{\ell}, a_{\ell}, O_{\ell+1})$ from consecutive frames sampled at the simulator control frequency, keeping simulator metadata only for offline analysis and for the label of $\mathcal{L}_{\mathrm{dist}}$. We remove corrupted frames, invalid robot states, reset fragments, terminal fragments without motion, and transitions that violate the platform action limits. Train, validation, and test splits are scene-disjoint (Table~\ref{tab:implementation_details}).

\begin{table}[!ht]
\centering
\caption{Implementation details: dataset scale, architecture, optimization, planner, and compute.}
\label{tab:implementation_details}
\setlength{\tabcolsep}{5pt}
\renewcommand{\arraystretch}{1.04}
\footnotesize
\begin{tabular}{p{2.3cm} p{10.8cm}}
\toprule
Item & Details \\
\midrule
Data &
EVT-Bench / Habitat 3.0; scene-disjoint splits.
Train scenes: 64 / 72; val: 8 / 9; test: 8 / 9.
Trajectories: 48{,}000 / 61{,}000.
One-step clips: 1.92M / 2.44M.
Filtering removes corrupted frames, invalid robot states, reset fragments, terminal-only collision fragments, and action-invalid transitions. \\
\midrule
Observation and action &
Observations sampled at the simulator control frequency; consecutive frames without temporal skipping; actions at the same rate. \\
\midrule
Model &
VAE~\citep{svd} (sd-vae-ft-ema; 34.2M-parameter encoder, 49.5M-parameter decoder), $224{\times}224$ views, latent $4{\times}28{\times}28$. Target-evidence branch input: the scalar $0.4\,H^{\mathrm{vis}} + 0.6\,(H^{\mathrm{area}})^{1/2}$ (the evidence head outputs the two components); conditioning on the current step only: one observation latent, no history; robot state: planar position and yaw; action $(v, \omega)$. Transition: 12 causal layers, each with target-evidence, robot (a pose sub-layer and an action sub-layer, giving $R_{\ell+1} = [x_{\ell+1}, W_a(a_{\ell})]$), and observation sub-layers (attention + FFN), hidden 512, 8 heads, FFN 2048; 153.6M parameters, 89\,GFLOPs per denoising pass. Evidence: GroundingDINO Swin-T (172.8M parameters), prompt ``humanoid'', views resized to $800{\times}800$, on the current observation and for training labels; evidence head $g_{\psi}$ (0.26M parameters) on imagined steps. \\
\midrule
Diffusion training &
Linear DDPM with fixed variance, 1000 train steps, $\varepsilon$-prediction on the next observation latent, 20 DDIM inference steps; AdamW, lr $8{\times}10^{-5}$, no weight decay, batch 16, 150 epochs, grad-norm clip 10, EMA 0.9999, bf16 mixed precision; 3 seeds; evidence-head loss weight $\lambda_{\mathrm{ev}}=1$ (joint training from the 12k-step transition); auxiliary distance-aware loss weight $\lambda_{\mathrm{dist}}=0.2$. \\
\midrule
Planning &
$T{=}10$, $N{=}128$, $K{=}16$, $M{=}4$, 20 DDIM steps; $\beta_{\mathrm{vis}}{=}\beta_{\mathrm{dist}}{=}\lambda_{\mathrm{valid}}{=}\lambda_{\mathrm{safe}}{=}1.0$; $\alpha{=}0.5$; $d_{\mathrm{safe}}{=}0.20$\,m; one stochastic rollout per candidate; target evidence read by $g_{\psi}$ at every imagined step. \\
\midrule
Compute &
Training: a single GPU per model, bf16 mixed precision, deterministic evaluation seeds.
Runtime: measured on one RTX A6000, and on an RTX 6000 Ada for the NWM-protocol trajectory times and the robot-budget planning step; per-component times, memory, and the NWM-protocol trajectory times are in Appendix~\ref{app:runtime} and Table~\ref{tab:runtime}.
Random seeds per main model: 3.
Evaluation episodes per model: 500 starting states $\times$ 32 candidate sequences for offline diagnostics, plus standard EVT-Bench / Habitat 3.0 splits for downstream metrics. \\
\bottomrule
\end{tabular}
\end{table}

\paragraph{Training and model selection.}
All compared world models are trained on the same transitions and split with the settings of Table~\ref{tab:implementation_details}; ablations keep the backbone, schedule, and planner fixed and change one component. Checkpoints are selected on the validation split by the mean of F and DRS, with ties (to three decimals) broken by lower CR, and every learned model is trained with three seeds.

\paragraph{Metrics.}
On EVT-Bench, success rate (SR), tracking rate (TR), and collision rate (CR) measure task completion, tracking continuity, and safety. On Habitat 3.0, following rate (F) measures how consistently the robot follows, distance-range success (DRS) the fraction of time within the following interval, CR safety, and episode success (ES) whether the episode ends with valid following. Under target loss, re-acquisition success (Re-acq.) and time to re-acquire (TTR) measure recovery, and following rate after re-acquisition (Post-F) and recovery episode success (Rec-ES) measure whether tracking resumes. Re-acquisition is scored against the original target, so following the distractor does not count. The offline diagnostics are paired interventions: 500 simulator states are saved from held-out episodes, and from each state the same 32 candidate action sequences are executed in the simulator and imagined by every model, so the simulated outcome of each candidate is its interventional outcome from that state. \emph{Rollout fidelity} compares predicted and simulated futures at horizons $\{1, 5, 10, 20, 40\}$ by latent error and by the AUROC of the evidence head's visibility reading against GroundingDINO on the simulated frame. \emph{Planning-value consistency} scores the 32 candidates of each state with Eq.~\eqref{eq:reward} and compares the ranking with the one given by their simulated outcomes by Spearman's $\rho$ and Kendall's $\tau$.

\paragraph{Baselines.}
The reactive policies (Uni-NaVid~\citep{uninavid}, TrackVLA~\citep{trackvla}, TrackVLA++ single-view~\citep{trackvlapp}, and OA-VAT~\citep{sun2026oavat} on EVT-Bench; the Habitat 3.0 baseline~\citep{h3} and SDA-S2~\citep{sda} on Habitat 3.0) act from their own inputs and training data and serve as system-level references. The adapted NWM~\citep{nwm} isolates the transition. We retrain NWM's CDiT-XL/2 once on our trajectories with our optimizer settings and replace its goal-image objective, which tracking cannot provide, with Eq.~\eqref{eq:reward}. It reads current evidence with GroundingDINO and imagined evidence with an evidence head of our architecture, trained with $\mathcal{L}_{\mathrm{ev}}$ on its own final-layer observation tokens, and it shares the observation interface, planning horizon, candidate budget, CEM iterations, sampling steps, and action bounds. What remains different is the transition: the action conditions every token of the adapted NWM and of the TrackVLA-style predictor, whereas our target-evidence branch is masked. Table~\ref{tab:baseline_alignment} summarizes what each method receives.

\begin{table}[!ht]
\centering
\caption{What each compared method receives at test time. ``Same as ours'' for the adapted NWM and the ablations means identical training data, optimizer settings, evidence read-out, planning objective (Eq.~\eqref{eq:reward}), and CEM budget ($T{=}10$, $N{=}128$, $K{=}16$, $M{=}4$, 20 DDIM steps); ablations differ from the full model only in the named component. $^{\dagger}$Run by us on EVT-Bench STT with the released pipeline, without the diffusion planner.}
\label{tab:baseline_alignment}
\setlength{\tabcolsep}{3.5pt}
\renewcommand{\arraystretch}{1.08}
\footnotesize
\begin{tabular}{>{\raggedright\arraybackslash}p{2.7cm} >{\raggedright\arraybackslash}p{1.7cm} >{\raggedright\arraybackslash}p{2.7cm} >{\raggedright\arraybackslash}p{2.3cm} >{\raggedright\arraybackslash}p{3.1cm}}
\toprule
Method & Type & Target evidence & Action selection & Model and compute \\
\midrule
Uni-NaVid, TrackVLA, TrackVLA++ & reactive VLA & learned, from RGB and instruction & one policy pass per step & own architecture and training data \\
OA-VAT$^{\dagger}$ & reactive tracker & own detector, tracker, and re-identification & PID control law & released weights, no planner \\
Habitat 3.0 baseline, SDA-S2 & reactive & own perception & one policy pass per step & own architecture and training data \\
Adapted NWM & world model, CEM & GroundingDINO on current view, evidence head on imagined views & Eq.~\eqref{eq:reward}, same budget & CDiT-XL/2, 1.01B, 321\,GFLOPs per pass; same as ours otherwise \\
Ablations (Table~\ref{tab:extra_ablation_breakdown}) & world model, CEM & GroundingDINO on current view, evidence head on imagined views & Eq.~\eqref{eq:reward}, same budget & full model with one component changed \\
CST-WM (ours) & world model, CEM & GroundingDINO on current view, evidence head on imagined views & Eq.~\eqref{eq:reward} & causal transition, 0.15B, 89\,GFLOPs per pass \\
\bottomrule
\end{tabular}
\end{table}

\paragraph{OA-VAT on EVT-Bench.}
OA-VAT~\citep{sun2026oavat} is evaluated on UnrealCV and a real drone. We run its released code on the 1{,}404 episodes of EVT-Bench STT: first-frame target search by prototype matching (YOLOE candidates and DINOv3 features), ORTrack tracking with its confidence-aware Kalman filter, prediction and re-acquisition modes, online prototype updates, and its PID control law. The diffusion planner checkpoint is not released, so we run the PID variant that OA-VAT reports in its own ablation. The reference image of the target is the first frame, with the simulator's box at $t=0$ standing in for the box a user would draw; the simulator box is not used afterwards. The control law uses EVT-Bench's image-space gains and set-point, since OA-VAT's drone gains are in UnrealCV units. Thresholds are those of the released evaluation script. Episodes are terminated and scored with EVT-Bench's own evaluation code.

\paragraph{Real-world protocol.}
Following TrackVLA~\citep{trackvla}, each method runs 10 trials per scenario on the same course. In \emph{occlusion}, the target walks behind bookshelves and around corridor corners; in \emph{distractor}, a second person in similar clothing crosses between the robot and the target; in \emph{fast motion}, the target walks quickly along a winding path. A trial succeeds if the robot follows the target to the end of the course without collision and without losing it for more than 6.8\,s. The tracking rate is the fraction of time the target is in view at 1--3\,m, measured from the recorded video. Following the distractor instead of the target is a failure. With 10 trials per scenario, the 95\% Wilson intervals of the success rates are 49--94\% for 8/10, 31--83\% for 6/10, and 17--69\% for 4/10; over all 30 trials they are 49--81\% for CST-WM (20/30) and 30--64\% for TrackVLA (14/30). To run on the robot, planning uses a smaller budget than in simulation, $T{=}4$, $N{=}24$, $K{=}8$, $M{=}2$, and 10 sampling steps, on an offboard RTX 6000 Ada: one planning step takes 1.4\,s (Table~\ref{tab:runtime_e2e}), and between planning steps the robot executes the last chosen action.

\paragraph{Evidence head for deployment.}
All simulation results, including Table~\ref{tab:readout_closedloop}, use the evidence head trained on EVT-Bench and Habitat 3.0 as described in Section~\ref{sec:causal-diffusion}. For the real-world trials only, the evidence head is further trained on 100k real frames containing people, taken from publicly available online videos, mixed with 100k frames collected in EVT-Bench and Habitat 3.0, with the transition frozen. Real frames are labeled by GroundingDINO as for $\mathcal{L}_{\mathrm{ev}}$, and their camera motion, which the videos do not record, is estimated by visual odometry and used as the action input of the forward pass.

\FloatBarrier
\section{Additional Experimental Results}
\label{app:more_results}

\subsection{Structural Diagnostics}
\label{app:structural}
These diagnostics complement Table~\ref{tab:structural}. For our model the action-dependence of $E_{\ell+1}$ is zero by construction, so its rows verify the implementation of the mask; the informative rows are those of models whose target representation may read the action, which show how much action information such a representation absorbs when it is allowed to. Table~\ref{tab:jacobian_distribution} breaks the Jacobian $|\partial E_{\ell+1}/\partial a_{\ell}|$ down by target visibility: the adapted NWM and the leaky variants depend on the action both in and out of view, and their dependence correlates negatively with tracking performance across episodes ($-0.36$ to $-0.53$). Table~\ref{causal} varies how deep the mask reaches: without a mask (A, single branch) the mean Jacobian is 0.19, a shallow mask (B) raises it to 0.33, and partial and deeper masks (C, D) lower it to 0.09 and 0.04. Table~\ref{tab:interventional_invariance} replaces $a_{\ell}$ with other actions from $\mathcal{A}_{\mathrm{valid}}$ at a fixed state and measures the spread of the resulting target representation. The robot-motion-only rows of Table~\ref{tab:structural} test what the mask does not enforce: with the humanoid held still, stability is one minus the normalized variation of $E_{\ell+1}$, which can change through $Z_{\ell}$ and $x_{\ell}$, and view sensitivity is the normalized variation of $\widehat{Z}$ under viewpoint change.

\begin{table}[!htbp]
\centering
\caption{Jacobian $|\partial E_{\ell+1}/\partial a_{\ell}|$ of the target-evidence representation over all rollout states and by target visibility, and its correlation with per-episode tracking performance. Best in \textbf{bold}, second \underline{underlined} (Jacobian columns).}
\label{tab:jacobian_distribution}
\setlength{\tabcolsep}{6pt}
\footnotesize
\begin{tabular}{lcccc}
\toprule
Method & All states$\downarrow$ & In-view$\downarrow$ & Out-of-view$\downarrow$ & Corr.\ with performance \\
\midrule
Adapted NWM & 0.112 & 0.096 & 0.131 & $-0.41$ \\
Leaky variant B & 0.331 & 0.304 & 0.357 & $-0.53$ \\
Leaky variant C & \underline{0.086} & \underline{0.071} & \underline{0.102} & $-0.36$ \\
Ours & \textbf{0.001} & \textbf{0.001} & \textbf{0.002} & $-0.04$ \\
\bottomrule
\end{tabular}
\end{table}

\begin{table}[!htbp]
\centering
\begin{minipage}[t]{0.44\linewidth}
\centering
\caption{Mean Jacobian $|\partial E_{\ell+1}/\partial a_{\ell}|$ for masks of increasing depth: A, none (single branch); B, shallow; C, partial; D, deeper.}
\label{causal}
\setlength{\tabcolsep}{4pt}
\footnotesize
\begin{tabular}{lccccc}
\toprule
Mask & A & B & C & D & Ours \\
\midrule
Jacobian & 0.19 & 0.33 & 0.09 & \underline{0.04} & \textbf{0.00} \\
\bottomrule
\end{tabular}
\end{minipage}\hfill
\begin{minipage}[t]{0.53\linewidth}
\centering
\caption{Target representation under action swap at a fixed state: variance across actions and mean pairwise $W_1$ distance.}
\label{tab:interventional_invariance}
\setlength{\tabcolsep}{4pt}
\footnotesize
\begin{tabular}{lcc}
\toprule
Method & Variance$\downarrow$ & Pairwise $W_1$$\downarrow$ \\
\midrule
Adapted NWM & \underline{0.084} & \underline{0.127} \\
Leaky variant & 0.146 & 0.214 \\
Ours & \textbf{0.006} & \textbf{0.011} \\
\bottomrule
\end{tabular}
\end{minipage}
\end{table}

\FloatBarrier
\subsection{Reading Imagined Target Evidence}
\label{app:evidence_readout}
These protocols belong to the read-out experiments of Sections~\ref{sec:offline} and~\ref{sec:ablation} (Tables~\ref{tab:evidence_readout} and~\ref{tab:readout_closedloop}).

\paragraph{Offline protocol.}
Table~\ref{tab:evidence_readout} uses 192 starts on held-out trajectories, half chosen among starts whose target visibility changes within 8 steps and half at random, so that the steps a planner must anticipate are well represented. From each start, a model rolls out 8 steps with 20 DDIM steps per step, EMA weights, and the logged actions and poses; the sampled latent becomes the next context, and the head's own reading, $0.4\,\widehat{H}^{\mathrm{vis}} + 0.6\,\widehat{H}^{\mathrm{area}}$ (the head outputs the root of the area), becomes the next evidence input. All read-outs use the same sampled latents, so they differ only in how they read them. Labels are GroundingDINO readings of the real frames, obtained as for $\mathcal{L}_{\mathrm{ev}}$; a frame counts as visible when its confidence exceeds 0.35. The separate latent head is a three-stage convolutional network (64/128/256 channels, global average pooling, two-layer MLP) trained on encoder latents of the training frames with the same labels and with additive latent noise, blur, shifts, and horizontal flips as augmentation. The full model and the w/o-masking variant are trained from the same transition with the same schedule; the latter lets the target-evidence branch attend to the action token and is otherwise identical. For $\Delta_{\mathrm{turn}}$, the angular velocity of every action is negated and the rollout is repeated with the same noise.

\paragraph{Steps at which visibility changes.}
Restricting the evaluation to the steps at which the target's visibility differs from the start, the steps a planner has to anticipate, the evidence head reaches a visibility AUROC of 0.467 with the mask and 0.446 without it, against 0.530 for detecting on the decoded view and 0.417 for the separate latent head; copying the current evidence is wrong on these steps by definition.

\paragraph{Closed-loop protocol.}
\label{app:readout_closedloop}
Table~\ref{tab:readout_closedloop} runs the Habitat 3.0 standard-tracking and target-loss episodes with the full transition and changes only the target evidence. The true-distance reference gives the planner the simulator's robot--target distance in place of the apparent-scale term. For the rendered-view reference, each CEM candidate is executed from a saved copy of the simulator state, the humanoid replays its recorded motion for the next $T$ steps, and GroundingDINO reads the rendered $224{\times}224$ views with the thresholds of $\mathcal{L}_{\mathrm{ev}}$. For decode-and-detect, the final denoised latent of every imagined step is decoded by $\mathcal{D}$ and read by the same detector. The frozen-transition head has the architecture of $g_{\psi}$ and is trained with $\mathcal{L}_{\mathrm{ev}}$ with all transition parameters fixed; the joint head follows the same schedule with all parameters updated, so the two differ only in whether $\mathcal{L}_{\mathrm{ev}}$ shapes the dynamics. Read-out time is for one imagined step, a batch of $N{=}128$ views, measured inside the closed-loop planner, excluding denoising, as the median over imagined steps after warm-up; Table~\ref{tab:runtime_components} times the head's forward pass alone.

\FloatBarrier
\subsection{Target-Evidence Signal and Seeds}
\label{app:signal}
\begin{figure}[!ht]
  \centering
  \includegraphics[width=0.6\linewidth]{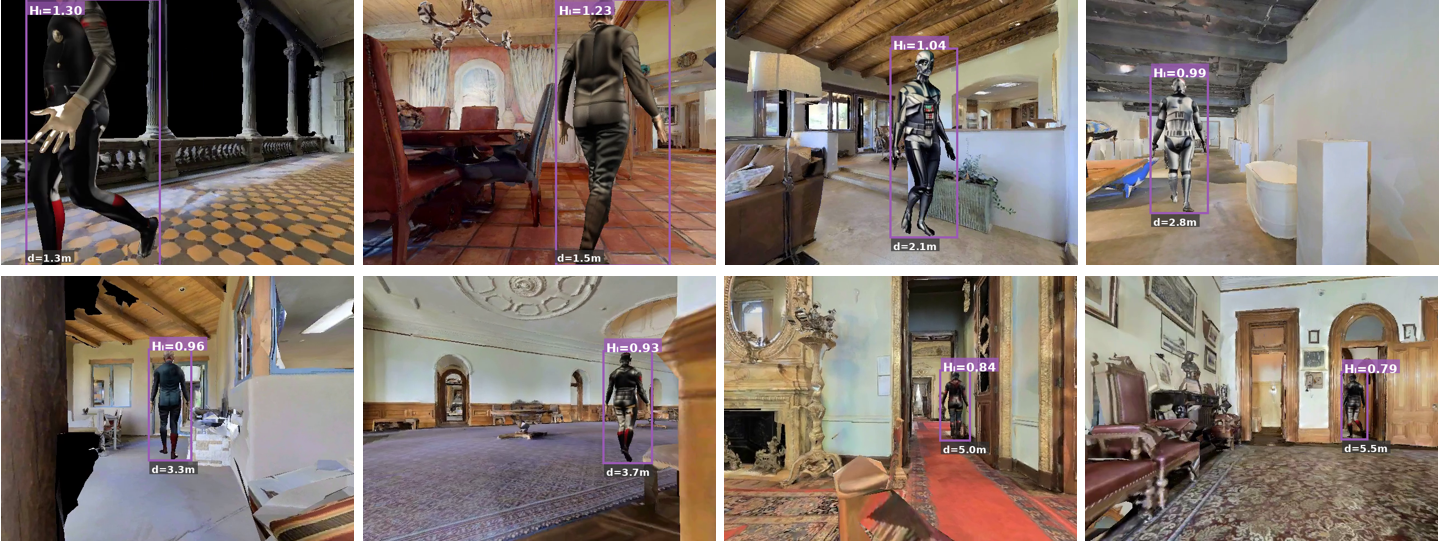}
  \caption{GroundingDINO target-evidence score against relative robot--human distance on EVT-Bench. These readings are the current evidence and the training labels of the evidence head.}
  \label{fig:grounddino}
\end{figure}

Figure~\ref{fig:grounddino} plots the detector's score against the relative distance on EVT-Bench: it tracks the distance over most of the following range and saturates only very close to the target, which is why the planner can use it as a distance signal. Table~\ref{tab:cross_humanoid} repeats the signal comparison across humanoids of different body scale, height, appearance, clothing, and gait: the distance-aware loss raises the Spearman correlation between the signal and the (sign-inverted) true distance from 0.71 to 0.83 and improves all four downstream metrics. Table~\ref{tab:seed_breakdown} shows the full model on Habitat 3.0 standard tracking for three seeds, which vary by at most 0.02, against a gap of 0.12 in F to the strongest baseline in Table~\ref{tab:main_quantitative}.

\begin{table}[!htbp]
\centering
\begin{minipage}[t]{0.66\linewidth}
\centering
\caption{Signal across humanoids of different appearance and gait on Habitat 3.0. $\rho$: Spearman correlation between the signal and the sign-inverted true distance.}
\label{tab:cross_humanoid}
\setlength{\tabcolsep}{4pt}
\footnotesize
\begin{tabular}{lccccc}
\toprule
Variant & $\rho$$\uparrow$ & F$\uparrow$ & DRS$\uparrow$ & Re-acq.$\uparrow$ & CR$\downarrow$ \\
\midrule
W/o distance-aware loss & 0.71 & 0.47 & 0.62 & 0.68 & 0.31 \\
With distance-aware loss & \textbf{0.83} & \textbf{0.52} & \textbf{0.69} & \textbf{0.75} & \textbf{0.28} \\
\bottomrule
\end{tabular}
\end{minipage}\hfill
\begin{minipage}[t]{0.3\linewidth}
\centering
\caption{Full model per seed, Habitat 3.0 standard tracking.}
\label{tab:seed_breakdown}
\setlength{\tabcolsep}{4pt}
\footnotesize
\begin{tabular}{lccc}
\toprule
Seed & F$\uparrow$ & DRS$\uparrow$ & CR$\downarrow$ \\
\midrule
1 & 0.52 & 0.69 & 0.28 \\
2 & \textbf{0.54} & \textbf{0.71} & \textbf{0.26} \\
3 & \underline{0.53} & \underline{0.70} & \underline{0.27} \\
\bottomrule
\end{tabular}
\end{minipage}
\end{table}

\FloatBarrier
\subsection{Runtime Details}
\label{app:runtime}

Unless marked Ada, times are measured on one NVIDIA RTX A6000 (48\,GB) in a workstation, as the median of repeated runs after warm-up, with the world model in bf16 autocast and eager mode (PyTorch 2.4) and GroundingDINO in fp32 as in its reference inference code (PyTorch 2.5); no quantization or distillation is used.

\paragraph{One planning step, from observation to action.}
A CEM planning step with $N$ candidates, horizon $T$, $M$ iterations, and $S$ sampling steps runs $MTS$ denoising passes, each over all $N$ candidates as one batch, in sequence: the $S$ passes of one imagined step, the $T$ steps of one rollout, and the $M$ iterations of CEM each depend on the previous one. The evidence head reads the $MT$ imagined steps from the tokens of the final pass of each step, GroundingDINO reads the current observation once, and the current frame is encoded once. Table~\ref{tab:runtime_e2e} gives the wall-clock time of the whole step for the three budgets, measured by running the full step, with the count of every stage; the detector and the encoder take 0.1\,s of it and the head reads under 0.02\,s, so the denoising passes account for the rest. Denoising compute per step is $MTSN \times 89$\,GFLOPs, the horizontal axis of Figure~\ref{fig:runtime_pareto}. In simulation the environment is stepped synchronously, so the planner's time does not affect the simulated results; on the robot, planning runs offboard with the robot budget, and the real-world protocol of Appendix~\ref{app:exp_setting} states the resulting planning rate.

\begin{table}[!ht]
\centering
\caption{Wall-clock time of one planning step from the received observation to the issued action, for the three budgets, on one RTX A6000 (median of full steps after warm-up); the robot budget is also given on the offboard RTX 6000 Ada of the real-world trials. Denoising passes run in sequence, each over the $N$ candidates as one batch; per-pass times are in Table~\ref{tab:runtime_components}.}
\label{tab:runtime_e2e}
\setlength{\tabcolsep}{4pt}
\footnotesize
\begin{tabular}{lcccc}
\toprule
Stage & Default & Fast & Robot & Robot, Ada \\
Budget $(T, N, M, S)$ & (10,128,4,20) & (6,64,2,10) & (4,24,2,10) & (4,24,2,10) \\
\midrule
GroundingDINO, current view & 0.09\,s & 0.09\,s & 0.09\,s & -- \\
VAE encoder, current frame & 7\,ms & 7\,ms & 7\,ms & -- \\
Denoising passes (sequential, batch $N$) & 800 & 120 & 80 & 80 \\
Evidence-head reads (batch $N$) & 40 & 12 & 8 & 8 \\
\midrule
Time per action (whole step) & 267\,s & 22.5\,s & 8.2\,s & 1.4\,s \\
Planning steps per second & 0.0038 & 0.044 & 0.12 & 0.71 \\
\bottomrule
\end{tabular}
\end{table}

\paragraph{Components and the NWM protocol.}
Table~\ref{tab:runtime_components} lists the components of a planning step at the candidate batch sizes of the three budgets; its per-pass times are isolated calls, and inside a full planning step a pass takes 3--8\% less (Table~\ref{tab:runtime_e2e} divided by its pass count). Table~\ref{tab:runtime} follows the NWM protocol~\citep{nwm}: one trajectory is simulated at batch size~1, and each state costs a VAE encoding of the 4-frame context, the diffusion sampler, and a VAE decoding, exactly as in the NWM inference code (our transition reads only the last frame of that context); the NWM CDiT-XL/2 backbone is timed with the same script on the same GPU, and our model is also timed on an RTX 6000 Ada, the GPU on which NWM reports its times. This protocol times the throughput of one trajectory, not a planning step. As in NWM, the 4-bit column divides the 6-step time by 4 without running quantization. NWM obtains its 6-step column by distillation; our 6-step column times a 6-step DDIM sampler. With \texttt{torch.compile}, a batch-1 denoising pass takes 24.7\,ms and the 6-step trajectory of Table~\ref{tab:runtime} 1.53\,s; at batch sizes of 64 and above the compiled model is not faster, so all planning-step times use eager mode.

\begin{table}[!ht]
\centering
\caption{Time (s) to simulate one trajectory at batch size~1 (NWM protocol): 16 states with 250 steps (Base), 8 states (+Skip), 6 sampling steps, and 4-bit quantization. Ada: RTX 6000 Ada. $^{\ast}$Estimated as in NWM: 6-step time divided by 4. $^{\dagger}$NWM distills to 6 steps; we time a 6-step sampler.}
\label{tab:runtime}
    \setlength{\tabcolsep}{2.4pt}
    \footnotesize
\begin{tabular}{lcccc}
    \toprule
    Model, GPU & Base & +Skip & +6 steps$^{\dagger}$ & +4-bit$^{\ast}$ \\
    \midrule
    NWM, Ada (reported) & 30.3 & 14.7 & 0.4 & 0.1 \\
    NWM, A6000 & 166.2 & 84.0 & 2.30 & 0.57 \\
    Ours, A6000 & 135.6 & 67.7 & 1.88 & 0.47 \\
    Ours, Ada & 23.4 & 12.3 & 0.34 & 0.08 \\
    \bottomrule
    \end{tabular}
\end{table}

\begin{table}[!htbp]
\centering
\caption{Per-component runtime and peak allocated GPU memory on one RTX A6000. Denoising passes are timed at the candidate batch sizes of the robot budget and of the fast, default, and large budgets of Figure~\ref{fig:runtime_pareto}; the NWM backbone is listed for reference. The evidence head reads imagined views during planning, GroundingDINO reads the current observation, and the decoder serves visualization and the decode-and-detect read-out of Table~\ref{tab:readout_closedloop}. $^{\ast}$Batch size~1 includes CPU resizing in the reference code.}
\label{tab:runtime_components}
\setlength{\tabcolsep}{4.5pt}
\footnotesize
\resizebox{\textwidth}{!}{%
\begin{tabular}{llccc}
\toprule
Component & Size & Batch & Time & Peak memory \\
\midrule
VAE encoder, 1 / 4 frames & 34.2M & 1 & 7 / 23\,ms & -- \\
Ours, one denoising pass & 153.6M, 89\,GFLOPs & 1 / 24 / 64 & 31 / 106 / 198\,ms & 1.2 / 1.6 / 2.4\,GB \\
 & & 128 / 192 & 362 / 572\,ms & 3.6 / 4.8\,GB \\
NWM CDiT-XL/2, one denoising pass & 1.01B, 321\,GFLOPs & 1 / 128 & 39 / 1017\,ms & -- \\
Evidence head $g_{\psi}$ & 0.26M & 24 / 64 / 128 / 192 & 0.21 / 0.20 / 0.20 / 0.27\,ms & -- \\
VAE decoder, per view & 49.5M & 1 / 64 & 15.6 / 10.6\,ms & -- \\
GroundingDINO Swin-T, per view$^{\ast}$ & 172.8M, $800{\times}800$ & 1 / 8 & 90 / 40\,ms & 1.0 / 3.5\,GB \\
 & & 32 & 38\,ms & 12.0\,GB \\
\bottomrule
\end{tabular}%
}
\end{table}

\FloatBarrier
\section{Limitations}
\label{app:limitations}

\paragraph{Proxy, not target state.} CST-WM is designed for planning-oriented embodied visual tracking rather than full target-state recovery. Its target evidence summarizes observability and apparent scale, so re-acquisition means that imagined views contain the target again within the planning horizon; the model does not represent the hidden target's trajectory, and when no candidate brings the target back within $T$ steps the planner acts on the validity and safety terms until new evidence arrives (Section~\ref{sec:planning}).

\paragraph{Detector dependence.} Target evidence is read by an evidence head trained on GroundingDINO labels, so planning inherits the detector's errors: miscalibrated confidence under partial visibility, weak responses for very small or distant targets, saturation of the area term at very close range (Figure~\ref{fig:grounddino}), and confusion between similar-looking humans. The model has no persistent notion of target identity: its evidence says that a humanoid the detector accepts is in view, so several candidate humans or a distractor that looks like the target can redirect the planner. The distractor-crossing condition of Table~\ref{tab:reacquisition} and the cross-humanoid evaluation (Table~\ref{tab:cross_humanoid}) cover part of these conditions.

\paragraph{Evaluation scope.} EVT-Bench results cover its single-target tracking task (STT); its distracted (DT) and ambiguous (AT) tracking tasks require the robot to keep the identity of one target among similar people, which is outside the scope of CST-WM.

\end{document}